\documentclass[sigplan,twocolumn,10pt]{acmart}
\AtBeginDocument{%
  }

\copyrightyear{2026}
\acmYear{2026}
\setcopyright{rightsretained}
\acmConference[EUROSYS '26]{21st European Conference on Computer Systems}{April 27--30, 2026}{Edinburgh, Scotland Uk}
\acmBooktitle{21st European Conference on Computer Systems (EUROSYS '26), April 27--30, 2026, Edinburgh, Scotland Uk}
\acmPrice{}
\acmDOI{10.1145/3767295.3769331}
\acmISBN{979-8-4007-2212-7/26/04}



\usepackage{microtype}
\usepackage{graphicx}
\usepackage{booktabs} 
\usepackage{tikz}
\usepackage{amsmath}
\usepackage{multirow}
\usepackage{fontawesome}
\usepackage{subcaption}
\usepackage{xspace}
\usepackage{booktabs}
\usepackage{comment}
\usepackage{float}
\usepackage{caption}
\usepackage{lipsum}
\usepackage[]{hyperref}
\usepackage{comment}
\usepackage{xspace}
\usepackage{wasysym}
\usepackage{float}
\usepackage{stfloats}
\usepackage{placeins}
\usepackage{enumitem}
\usepackage{xcolor}
\usepackage{booktabs}
\PassOptionsToPackage{table}{xcolor}
\PassOptionsToPackage{x11names}{xcolor}
\usepackage{xcolor}
\usepackage{colortbl}
\usepackage{makecell}
\usepackage{nicematrix}
\usepackage[ruled,linesnumbered]{algorithm2e}
\usepackage[export]{adjustbox}
\usepackage{float}
\usepackage[noend]{algpseudocode}
\usepackage{tabularx} 

\usepackage{amssymb}

\usepackage{newtxmath}
\usepackage{pifont} 

\definecolor{darkergreen}{RGB}{0,100,0} 
\definecolor{darkerred}{RGB}{139,0,0} 
\definecolor{lightergreen}{RGB}{34,139,34} 
\definecolor{lighterred}{RGB}{205,92,92} 
\definecolor{mydarkgreen}{HTML}{2E5E17}

\SetCommentSty{textit}

\SetCommentSty{mygreen}

\usepackage{cleveref}
\crefformat{section}{\S#2#1#3} 
\crefformat{subsection}{\S#2#1#3}
\crefformat{subsubsection}{\S#2#1#3}

\usepackage{makecell}

\usepackage{xcolor} 
\usepackage{listings} 

\usepackage{pifont}
\definecolor{dmlgreen}    {RGB}{51,  160,  44}
\definecolor{dmlblue}     {RGB}{31,  120, 180}
\definecolor{dmlred}      {RGB}{202,   0,  32}
\definecolor{brown}       {RGB}{139,  69,  19}

\definecolor{deepblue}{rgb}{0,0,0.5}
\definecolor{deepred}{rgb}{0.6,0,0}
\definecolor{deepgreen}{rgb}{0,0.5,0}
\definecolor{mauve}{rgb}{0.58,0,0.82}
\definecolor{light-gray}{gray}{0.96}
\definecolor{aliceblue}{rgb}{0.94, 0.97, 1.0}
\definecolor{OliveGreen}{rgb}{0.33, 0.42, 0.18}

\DeclareMathOperator*{\argmin}{arg\,min}

\usepackage[utf8]{inputenc}
\DeclareFixedFont{\ttb}{T1}{txtt}{bx}{n}{8} 
\DeclareFixedFont{\ttm}{T1}{txtt}{m}{n}{8}  

\usepackage{xargs}                      
\usepackage[colorinlistoftodos,prependcaption,textsize=small]{todonotes}
\newcommandx{\note}[2][1=]{\todo[linecolor=OliveGreen,backgroundcolor=OliveGreen!25,bordercolor=OliveGreen,#1]{#2}}
\newcommandx{\unsure}[2][1=]{\todo[linecolor=red,backgroundcolor=red!25,bordercolor=red,#1]{#2}}
\newcommandx{\improve}[2][1=]{\todo[linecolor=orange,backgroundcolor=orange!25,bordercolor=orange,#1]{#2}}
\newcommandx{\change}[2][1=]{\todo[linecolor=Plum,backgroundcolor=Plum!25,bordercolor=Plum,#1]{#2}}
\newcommandx{\fix}[2][1=]{\todo[linecolor=blue,backgroundcolor=blue!25,bordercolor=blue,#1]{#2}}

\lstnewenvironment{env_python}[2]
{
\lstset{
  language=Python,
  backgroundcolor=\color{light-gray},
  otherkeywords={self},        
  emph={to,tile,reuse_at,reorder,unroll,split,vectorize,quantize},          
  emphstyle=\footnotesize\color{dmlred},  
  belowcaptionskip=0.7\baselineskip,
  aboveskip=-2mm,
  belowskip=5mm,
  showstringspaces=false,
  columns=flexible,
  escapechar=@,
  basicstyle={\linespread{0.99}\fontencoding{T1}\footnotesize\fontfamily{lmtt}\fontseries{m}\selectfont},
  numbers={left},
  xleftmargin={2em},%
  framesep=5pt, 
  framexleftmargin={1em},
  numberstyle=\tiny\color{gray},
  keywordstyle=\color{blue},
  commentstyle=\color{deepgreen},
  stringstyle=\color{mauve},
  breaklines=true,
  breakatwhitespace=true,
  tabsize=3,
  frame=tb, 
  caption= #1, 
  label= #2
}
  \vspace{\baselineskip}
}
{}

\newcommand{\lstbg}[3][0pt]{{\fboxsep#1\colorbox{#2}{\strut #3}}}
\lstdefinelanguage{diff}{
  backgroundcolor=\color{aliceblue},           
  emph={},          
  emphstyle=\small\color{dmlred},  
  belowcaptionskip=0.7\baselineskip,
  aboveskip=0mm,
  belowskip=3mm,
  showstringspaces=false,
  columns=flexible,
  basicstyle={\linespread{1.1}\fontencoding{T1}\scriptsize\fontfamily{lmtt}\fontseries{m}\selectfont},
  numbers={left},
  xleftmargin={2em},%
  breaklines=true,
  breakatwhitespace=true,
  tabsize=3,
  morecomment=[f][\lstbg{red!20}]-,
  morecomment=[f][\lstbg{green!20}]+,
  morecomment=[f][\textit]{@@},
}

\makeatletter
\let\old@lstKV@SwitchCases\lstKV@SwitchCases
\def\lstKV@SwitchCases#1#2#3{}
\makeatother
\usepackage{lstlinebgrd}
\makeatletter
\let\lstKV@SwitchCases\old@lstKV@SwitchCases

\lst@Key{numbers}{none}{%
    \def\lst@PlaceNumber{\lst@linebgrd}%
    \lstKV@SwitchCases{#1}%
    {none:\\%
     left:\def\lst@PlaceNumber{\llap{\normalfont
                \lst@numberstyle{\thelstnumber}\kern\lst@numbersep}\lst@linebgrd}\\%
     right:\def\lst@PlaceNumber{\rlap{\normalfont
                \kern\linewidth \kern\lst@numbersep
                \lst@numberstyle{\thelstnumber}}\lst@linebgrd}%
    }{\PackageError{Listings}{Numbers #1 unknown}\@ehc}}
\makeatother

\newcommand{\myparagraph}[1]{\vspace{1pt}\noindent\textbf{{#1}.\;}}
\newcommand{\filledone}{\ding{182}\xspace}
\newcommand{\filledtwo}{\ding{183}\xspace}
\newcommand{\filledthree}{\ding{184}\xspace}
\newcommand{\filledfour}{\ding{185}\xspace}
\newcommand{\filledfive}{\ding{186}\xspace}

\definecolor{customcolor}{HTML}{f9ae78}
\definecolor{customcolor2}{HTML}{3d5c6f}

\usepackage{xcolor}
\usepackage{listings}
\lstdefinestyle{mystyle}{
    basicstyle=\ttfamily\small,      
    numbers=left,                    
      numberstyle=\color{gray},   
    breakautoindent=false,
    breakindent=10pt,
    escapechar=|,
}
\usepackage{tikz}
\usetikzlibrary{patterns}
\usepackage[precision=2, unit=mm]{lengthconvert}

\newcommand{\Name}{LoRAFusion\xspace}

\begin{document}

\title{\Name: Efficient LoRA Fine-Tuning for LLMs}
\renewcommand{\shorttitle}{\Name: Efficient LoRA Fine-Tuning for LLMs}

\author{Zhanda Zhu}
\affiliation{%
  \institution{University of Toronto, Vector Institute, NVIDIA}
  \country{}
  }
\email{zhanda.zhu@mail.utoronto.ca}

\author{Qidong Su}
\affiliation{%
  \institution{University of Toronto, Vector Institute, NVIDIA}
  \country{}
  }
\email{qdsu@cs.toronto.edu}

\author{Yaoyao Ding}
\affiliation{%
  \institution{University of Toronto, Vector Institute, NVIDIA}
  \country{ }
  }
\email{dingyaoyao.cs@gmail.com}

\author{Kevin Song}
\affiliation{%
  \institution{University of Toronto, Vector Institute, NVIDIA}
  \country{}
  }
\email{xinyang.song@utoronto.ca}

\author{Shang Wang}
\affiliation{%
  \institution{University of Toronto, Vector Institute, NVIDIA}
  \country{}
  }
\email{wangsh46@cs.toronto.edu}

\author{Gennady Pekhimenko}
\affiliation{%
  \institution{University of Toronto, Vector Institute, NVIDIA}
  \country{}
  }
\email{pekhimenko@cs.toronto.edu}


\begin{abstract}
\sloppy
Low-Rank Adaptation (LoRA) has become the leading Parameter-Efficient Fine-Tuning (PEFT) method for Large Language Models (LLMs), as it significantly reduces GPU memory usage while maintaining competitive fine-tuned model quality on downstream tasks.
However, existing LLM LoRA fine-tuning systems mainly reuse optimizations from traditional full-model fine-tuning, and therefore cannot take full advantage of LoRA's unique characteristics.
We identify two key inefficiencies of existing works. 
First, existing LoRA fine-tuning systems incur substantial runtime overhead due to redundant memory accesses on large activation tensors.
Second, they miss the opportunity to concurrently fine-tune multiple independent LoRA adapters that share the same base model on the same set of GPUs. This leads to missed performance gains such as reduced pipeline bubbles, better communication overlap, and improved GPU load balance.

To address these issues, we introduce \Name, an efficient LoRA fine-tuning system for LLMs.
At the kernel level, we propose a graph-splitting method that fuses memory-bound operations. This design eliminates unnecessary memory accesses and preserves the performance of compute-bound GEMMs without incurring the cost of recomputation or synchronization.
At the scheduling level, \Name introduces an adaptive batching algorithm for multi-job fine-tuning. It first splits LoRA adapters into groups to intentionally stagger batch execution across jobs, and then solves a bin-packing problem within each group to generate balanced, dependency-aware microbatches.
\Name achieves up to $1.96\times$ ($1.47\times$ on average) end-to-end speedup compared to Megatron-LM, and up to $1.46\times$ ($1.29\times$ on average) improvement over mLoRA, the state-of-the-art multi-LoRA fine-tuning system.
Our fused kernel achieves up to $1.39\times$ ($1.27\times$ on average) kernel performance improvement and can directly serve as a plug-and-play replacement in existing LoRA systems.
We open-source \Name at \url{https://github.com/CentML/lorafusion}.
\end{abstract}

\maketitle

\section{Introduction}
\label{sec:introduction}

Pre-trained Large Language Models (LLMs), such as GPT~\cite{achiam2023gpt} and LLaMa~\cite{llama}, have demonstrated strong capabilities across diverse tasks, including text generation~\cite{brown2020gpt3,zhao2023survey}, question answering~\cite{lewis2020retrieval, guu2020retrieval}, and code generation~\cite{chen2021evaluating-code,nijkamp2022codegen}. 
To adapt these models to personalized or domain-specific tasks, \textit{fine-tuning} on pre-trained LLMs is typically performed.
Such adaptation is essential for scenarios like biomedical analysis~\cite{zhang2024generalist, tian2024opportunities}, personalized chatbot interactions~\cite{wang2022self-instruct}, or specialized customer support~\cite{yun2023fine-tuning-dialogue-summarization}. 
However, traditional full-model fine-tuning, where all model parameters are learned, requires substantial hardware resources such as multiple nodes, each node equipped with multiple flagship GPUs (e.g., NVIDIA B200 GPUs~\cite{B200}) and interconnected by high-speed links (e.g., NVLink~\cite{nvlink} and InfiniBand~\cite{infiniband}).  
For instance, full-model fine-tuning of LLaMa-3.1-70B~\cite{llama3.1} requires approximately 1120GB of GPU memory for model states alone (parameters, gradients, optimizer states), making this approach prohibitively expensive for practical applications~\cite{llama3.1-training-cost}.

\begin{sloppypar}
To mitigate the substantial hardware requirements, Parameter-Efficient Fine-Tuning (PEFT) methods~\cite{hu2022lora,li2021prefix-tuning,lester2021prompt-tuning,kopiczko2023vera,liu2024dora,dettmers2023qlora,li2023loftq,ding2023sparse-low-rank-adaptation,houlsby2019peft-learning-nlp,lin2020exploring,ruckle2020adapterdrop,zhao2024apt,zhang2025memory-efficient-lora} have emerged. These approaches significantly reduce resource usage by keeping pre-trained LLM parameters \textit{frozen} (not updated during fine-tuning) and selectively updating only a small set of injected trainable parameters, known as \textit{adapters}.
Among these methods, Low-Rank Adaptation (LoRA)~\cite{hu2022lora} and its variants~\cite{liu2024dora,dettmers2023qlora,li2023loftq,ding2023sparse-low-rank-adaptation} are particularly popular due to their simplicity and effectiveness. 
%
LoRA freezes the model's pre-trained weights and adds a residual branch composed of two trainable low-rank linear layers: a down-projection from the input dimension $k$ to a smaller rank $r$, followed by an up-projection back to the output dimension $n$, where $r \ll min(n, k)$. 
This low-rank branch is added to the output of the original frozen layer, enabling task-specific adaptation without modifying the base model (see Figure~\ref{fig:background-lora-illustration} and Section~\ref{subsec:background-llm-lora-finetuning}).
For example, fine-tuning LLaMa-3.1-70B~\cite{llama3.1} using LoRA adapters with a low-rank dimension of 16 introduces only 0.29\% additional parameters, reducing GPU memory usage to 142GB while preserving strong model quality~\cite{dettmers2023qlora,zhang2025memory-efficient-lora}.
\end{sloppypar}

Due to these substantial resource savings, LoRA fine-tuning has become widely adopted on both cloud platforms~\cite{openai-fine-tuning,together-ai-fine-tuning,runpod-fine-tuning,fireworks-fine-tuning} and local environments~\cite{zheng2024llamafactory,llama-cookbook}.
To realistically apply LoRA in practice and fine-tune high-quality adapters, users often run multiple jobs in parallel. These jobs may explore different hyperparameter settings~\cite{tribes2023hyperparameter-instruction-tuning} or continuously adapt models to evolving datasets and user needs~\cite{jindal2024balancing,ibrahim2024simple-continual-pre-training}.
As a result, fine-tuning throughput, measured as the number of training samples processed per second, has become a key metric for reducing both cost and overall training time.

However, despite significant algorithmic advances in LoRA-based approaches~\cite{dettmers2023qlora,kopiczko2023vera,liu2024dora} and explorations in serving scenarios~\cite{zhou2022pets,chen2024punica,sheng2023s-lora,wu2024dlora}, existing LoRA fine-tuning systems for LLMs largely reuse optimizations designed for full-model fine-tuning.
Specifically, systems like PEFT Library~\cite{peft}, LLaMA-Factory~\cite{zheng2024llamafactory}, and llama-cookbook~\cite{llama-cookbook} typically rely on a subset of parallelization methods such as Fully Sharded Data Parallelism (FSDP)~\cite{pytorch-fsdp,zero}, Tensor Parallelism (TP)~\cite{megatron-1,flexflow}, or Pipeline Parallelism (PP)~\cite{gpipe,pipedream,terapipe} to fit the training into GPU memory and achieve efficiency with multiple GPUs.
While such techniques are still useful, our analysis reveals they do not sufficiently address the unique characteristics of LoRA fine-tuning, causing significant inefficiencies.

The first limitation is the high runtime overhead introduced by LoRA adapters. 
LoRA adapters add less than 1\% parameters compared to the full model and are thus expected to incur minimal computation and memory overhead. 
However, our profiling shows that applying LoRA reduces training throughput by approximately 40\% compared to the original frozen model. 
This overhead comes from increased memory traffic: the small LoRA projection layers are memory-bandwidth-bound due to their low-rank dimensions, and the operations in the adapter repeatedly load and store large activation tensors. 
These factors together increase global memory access by up to $2.64\times$, as detailed in Section~\ref{sec:motivation-lora-overhead}.

The second limitation is the lack of efficient support for joint fine-tuning across multiple LoRA adapters. 
Typically, each adapter is fine-tuned independently, even if they share the same base model. 
Since LoRA adapters are lightweight and add minimal memory footprint pressure, multiple jobs can be combined and run jointly on the same GPUs. 
We refer to this strategy as \textit{multi-LoRA fine-tuning}~\cite{ye2023mlora}, which is highly practical for hyperparameter tuning~\cite{tribes2023hyperparameter-instruction-tuning} and multi-tenant cloud services~\cite{fireworks-fine-tuning,together-ai-fine-tuning}.
Although multi-LoRA optimization~\cite{chen2024punica,sheng2023s-lora,wu2024dlora} has been widely used in LLM serving, the motivation in our setting is entirely different. 
Serving systems batch requests to increase arithmetic intensity during autoregressive single-token decoding~\cite{chen2024punica,sheng2023s-lora}, whereas fine-tuning already processes full sequences with sufficient arithmetic intensity.
Instead, the primary benefit of multi-LoRA fine-tuning comes from mitigating the overhead of distributed parallelism~\cite{ye2023mlora}.

\begin{sloppypar}
While existing multi-LoRA fine-tuning systems like mLoRA~\cite{ye2023mlora} can reduce pipeline bubbles by filling them with independent groups of samples from different adapters, we observe that this approach is incomplete and still suffers from significant inefficiencies.
First, it still relies on generic LoRA kernels that are bottlenecked by redundant memory accesses to large activation tensors as discussed previously, and batching more samples does not help.
Second, it fails to address the load imbalance across GPUs.
Realistic fine-tuning workloads often contain samples with variable sequence lengths, leading to imbalanced work across GPUs.
This imbalance creates idle time in data-parallel replicas and increases pipeline bubbles from poorly aligned microbatches.
As we analyze in Section~\ref{sec:motivation-multi-lora-oversight}, strategically grouping and scheduling samples across multiple jobs is critical to mitigate this imbalance and unlock further efficiency gains.
\end{sloppypar}

Based on these insights, we argue that an efficient LoRA fine-tuning system must both reduce the runtime overhead of LoRA adapters and leverage multi-LoRA optimization opportunities to reduce distributed training overhead.
To meet these requirements, we propose \textit{\Name}, an efficient multi-level fusion system tailored specifically for LoRA fine-tuning of LLMs.
At the kernel level, our key insight is that most of LoRA's overhead comes from memory-bandwidth-bound operations on large activation tensors.
We address this by splitting the computation graph at the point where the tensor size shrinks to the low-rank dimension $r$.
This allows us to fuse memory-bandwidth-bound operations without recomputation or synchronization, while preserving the optimal performance of compute-bound matrix multiplications. 
This design leads to our \texttt{FusedLoRA} and \texttt{FusedMultiLoRA} kernels.
At the scheduling level, \Name enables concurrent fine-tuning of multiple LoRA adapters that share the same base model by batching samples across jobs.
Compared to the existing multi-LoRA systems, this further improves system throughput by reducing pipeline bubbles and balancing GPU workloads.
The challenge is that such batching must preserve execution dependencies between global batches, especially under pipeline parallelism.
To address this, we first group adapters in a way that creates natural gaps between their batches, then solve a bin-packing problem within each group to construct balanced, dependency-safe microbatches using a combination of Mixed Integer Linear Programming (MILP) and greedy heuristics.

We implement \Name on top of Megatron-LM~\cite{megatron-1}, a state-of-the-art distributed training framework, to support efficient parallelism and scalability.
We then extensively evaluate \Name on a wide range of LLMs, including LLaMa-3.1-8B, Qwen-2.5-32B, and LLaMa-3.1-70B, across NVIDIA H100 and L40S GPUs with various realistic datasets. 
Results show that \Name achieves up to $1.96\times$ ($1.47\times$ on average) end-to-end speedup compared to Megatron-LM with FSDP and PP, and up to $1.46\times$ ($1.29\times$ on average) improvement over mLoRA~\cite{ye2023mlora}, the state-of-the-art multi-LoRA fine-tuning system.
Additionally, our fused kernel alone achieves up to $1.39\times$ ($1.27\times$ on average) speedup compared to the default LoRA implementation on NVIDIA H100 GPUs, and can directly serve as a plug-in replacement in existing systems, offering immediate benefits to the broader community.

Overall, this paper makes the following contributions:

\begin{figure}[t]
    \centering
    \includegraphics[width=0.99\linewidth]{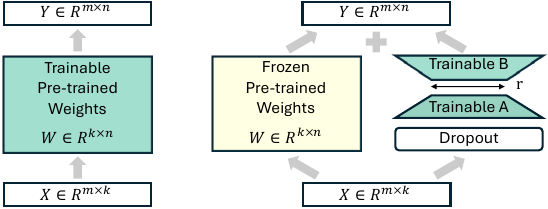}
    \hspace{0.01\linewidth} (a) Full Fine-tuning
    \hspace{0.125\linewidth} (b) LoRA Fine-tuning
    \hspace{0.125\linewidth}
    \caption{Comparison between traditional full model fine-tuning and LoRA fine-tuning.}
    \Description{Comparison between traditional full model fine-tuning and LoRA fine-tuning. Traditional fine-tuning updates all model parameters, while LoRA fine-tuning only updates lightweight adapter matrices.}
    \label{fig:background-lora-illustration}
\end{figure}

\begin{sloppypar}
\begin{itemize}
    \item We identify two key limitations in existing LoRA fine-tuning systems: high runtime overhead from redundant memory access, and missed opportunities for optimizing multi-job training. To address both, we propose \Name, a multi-level fusion system for accelerating LLM fine-tuning on modern GPU clusters.
    \item We propose a horizontal fusion strategy that reduces redundant memory access without disrupting compute-bound performance. We also design a scheduling algorithm that groups adapters across fine-tuning jobs and batches their samples to improve GPU load balance and reduce pipeline overhead.
    \item We evaluate \Name across diverse LLMs, datasets, and GPU platforms, showing significant improvements over existing LoRA fine-tuning systems. Our fused kernel also provides strong standalone gains and can be directly integrated into more general LoRA systems.
\end{itemize}
\end{sloppypar}

\normalsize
\section{Background}
\label{sec:background}

We first provide an introduction to LoRA fine-tuning (§\ref{subsec:background-llm-lora-finetuning}), and then present an overview of current system-level optimizations for fine-tuning (§\ref{subsec:background-system-optimization}).

\begin{table}[t]
    \centering
    \caption{Notation for the LoRA fine-tuning in this paper.}
     \begin{tabularx}{\linewidth}{c X}
     \toprule
     Symbol & Description \\
     \midrule
     $r$ & LoRA rank \\
     $m$ & Number of tokens (batch size $\times$ seq length) \\
     $k, n$ & Input/Output dimension of the weight matrix \\
     \midrule
     $W$ & Base model weights. Size: $(k, n)$. \\
     $A$ & First LoRA weights. Size: $(k, r)$ \\
     $B$ & Second LoRA weights. Size: $(r, n)$ \\
     $X$ & Input tensor. Size: $(m, k)$ \\
     $\widehat{X}$ & Input tensor after dropout. Size: $(m, k)$ \\
     $Y$ & Output tensor. Size: $(m, n)$ \\
     \bottomrule
     \end{tabularx}
    \label{tab:background-tab-1-notation}
\end{table}

\subsection{LLM LoRA Fine-tuning}
\label{subsec:background-llm-lora-finetuning}

Training an LLM from scratch demands substantial amounts of data, millions of GPU hours, and significant costs~\cite{llama3.1-training-cost}.
Fine-tuning pre-trained LLMs, such as LLaMa~\cite{llama3,llama3.1} and Qwen~\cite{bai2023qwen,yang2024qwen2-5} is thus more practical.
Fine-tuning preserves pre-trained capabilities while adapting the model to specialized downstream tasks~\cite{devlin2018bert}.
While fine-tuning reduces the data needs and training iterations, traditional full-model fine-tuning still requires similarly large GPU memory, due to the vast number of trainable parameters.

PEFT methods~\cite{hu2022lora,li2021prefix-tuning,lester2021prompt-tuning} address the memory requirements by freezing pre-trained parameters (keeping them non-updatable) and only training a small set of newly introduced parameters called \textit{adapters}.
Among these methods, LoRA~\cite{hu2022lora} is currently most widely used due to its simplicity and effectiveness. As illustrated in Figure~\ref{fig:background-lora-illustration}, LoRA injects two small trainable matrices alongside each pretrained weight matrix.
Formally, for a pretrained weight matrix $W \in \mathbb{R}^{k \times n}$, LoRA introduces two low-rank matrices $A \in \mathbb{R}^{k \times r}$ and $B \in \mathbb{R}^{r \times n}$, where LoRA rank $r \ll k,n$, combined as:
\begin{equation}
Y = X W + \alpha S B = X W + \alpha (\widehat{X} A) B
\end{equation}
where $\widehat{X} \in \mathbb{R}^{m \times k}$ is the input after dropout of $X \in \mathbb{R}^{m \times k}$, $S=\widehat{X} A$ is the intermediate result, $m$ is the number of aggregated tokens (batch size multiplied by sequence length), and $\alpha$ is a constant scalar for scaling.
Table~\ref{tab:background-tab-1-notation} summarizes key notations.

When fine-tuning LLMs, LoRA is applied to linear layers, replacing each original layer of dimensions $k \times n$ with a LoRA-equipped version. 
Assuming half-precision training with full-precision optimizer~\cite{megatron-1,zero}, memory usage of model states per linear layer decreases from $16nk$ bytes to $2nk + 32r(n+k)$ bytes. 
Since $r$ is much smaller than $n$ and $k$, the memory footprint of trainable parameters, gradients, and optimizer states is negligible compared to the pre-trained weights.
For instance, with $n$=$k$=4096 and $r$=16, LoRA parameters account for just about 0.39\% of the pre-trained weights. 
This dramatically reduces the memory required for gradients and optimizer states, decreasing memory demands by nearly $8\times$ compared to full-model fine-tuning.

\subsection{System Optimization for Fine-tuning}
\label{subsec:background-system-optimization}

Despite algorithmic advancements in LoRA~\cite{dettmers2023qlora,kopiczko2023vera,liu2024dora}, and optimizations for LoRA serving~\cite{chen2024punica,sheng2023s-lora,wu2024dlora}, existing system optimizations for LoRA fine-tuning primarily reuse techniques originally developed for LLM pre-training, including parallelization~\cite{pytorch-distributed,tensorflow,zero,pytorch-fsdp,megatron-1,zheng2022alpa}, kernel fusion~\cite{tvm,hidet,ansel2024pytorch-compile,dao2022flashattention,hsu2024liger,tillet2019triton,ashari2015optimizing}, and data packing~\cite{kundu2024hf-packing-with-flash-attention,wang2024packing}.

\myparagraph{Parallelization}
Parallelization partitions the computation and memory usage of the training process across multiple GPUs.
Data parallelism (DP)~\cite{pytorch-distributed,tensorflow} replicates the model across GPUs and partitions data batches. 
Fully Sharded Data Parallelism (FSDP or ZeRO-3)~\cite{zero,pytorch-fsdp} partitions model states and communicates them only when necessary, significantly reducing memory usage.
Tensor Parallelism (TP)~\cite{megatron-1, megatron-2} splits linear layers across GPUs and merges partial results via communication. 
Pipeline Parallelism (PP)~\cite{gpipe,pipedream,dapple} divides the model into sequential stages executed in a pipeline manner, reducing communication overhead but potentially introducing idle time (pipeline bubbles).

\myparagraph{Kernel Fusion}
Kernel fusion improves efficiency by merging multiple operations into fewer GPU kernels, reducing memory transfers and kernel launch overhead~\cite{tvm,hidet,ansel2024pytorch-compile,dao2022flashattention,hsu2024liger,tillet2019triton,ashari2015optimizing}. 
Recent techniques like Flash-Attention~\cite{dao2022flashattention} and element-wise kernel fusion~\cite{tillet2019triton,hsu2024liger} significantly improve performance by fusing frequently used operations in LLMs.
For LoRA, specialized kernels~\cite{chen2024punica,sheng2023s-lora} for multi-adapter serving are proposed, but these kernels do not directly improve fine-tuning throughput, as we discuss in Section~\ref{sec:motivation}.

\begin{figure}[t]
    \centering
    \includegraphics[width=0.995\linewidth]{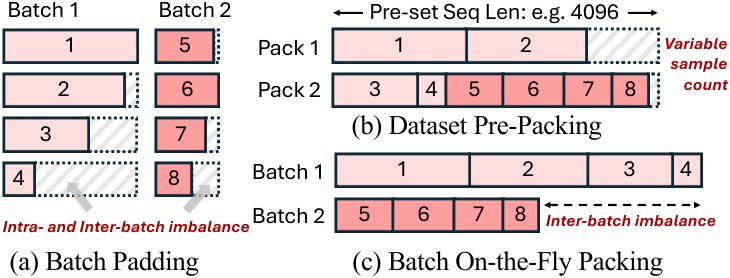}
    \caption{Comparison of (a) traditional batch padding, (b) dataset pre-packing, and (c) batch on-the-fly packing methods for LoRA fine-tuning of LLMs.}
    \Description{Padding and packing for LoRA fine-tuning.}
    \label{fig:background-padding-and-packing}
\end{figure}

\myparagraph{On-the-fly Data Packing}
When fine-tuning LLMs, training data often consists of token sequences of variable lengths. 
As depicted in Figure~\ref{fig:background-padding-and-packing}(a), traditional padding aligns shorter samples with padding tokens, causing wasted computations. 
Figure~\ref{fig:background-padding-and-packing}(b) shows the dataset pre-packing, which forms fixed-length batches in advance, but introduces variable sample counts per batch, potentially affecting training stability and randomness if not properly handled~\cite{kundu2024hf-packing-with-flash-attention}.
In contrast, on-the-fly packing (Figure~\ref{fig:background-padding-and-packing}(c)) dynamically concatenates samples within each batch, avoiding wasted computations while maintaining deterministic training samples per batch. Given its effectiveness and wide usage, we adopt on-the-fly packing throughout our work.

\normalsize

\section{Motivation}
\label{sec:motivation}

We identify two key limitations in existing LoRA fine-tuning systems: significant runtime overhead from redundant memory access (§\ref{sec:motivation-lora-overhead}), and missed optimization opportunities in multi-job training scenarios (§\ref{sec:motivation-multi-lora-oversight}).

\subsection{Significant Runtime Overhead of LoRA Modules}
\label{sec:motivation-lora-overhead}

\begin{figure}[t]
    \centering

    \begin{minipage}[c]{0.49\linewidth}
        \centering
        \includegraphics[width=\linewidth]{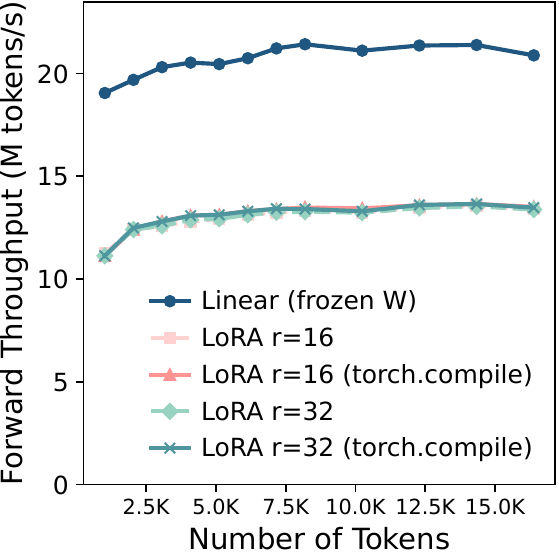}
        {\small (a) Forward pass throughput}
    \end{minipage}
    \hfill
    \begin{minipage}[c]{0.49\linewidth}
        \centering
        \includegraphics[width=\linewidth]{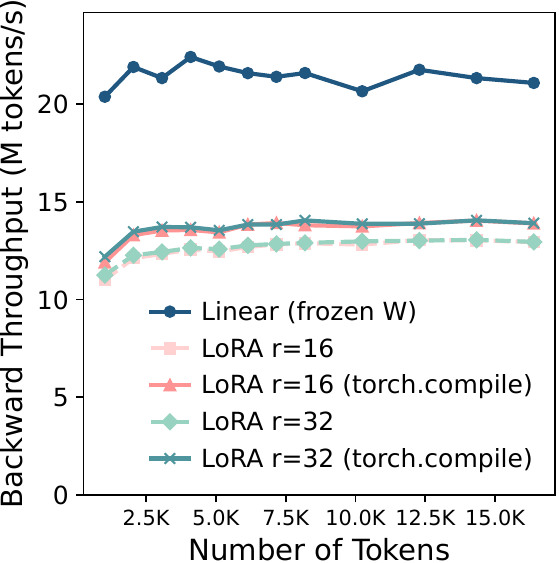}
        {\small (b) Backward pass throughput}
    \end{minipage}

    \caption{Throughput comparison of the frozen linear layer ($n$=$k$=4096) vs. the corresponding LoRA linear layer with different numbers of tokens and ranks.}
    \label{fig:motivation-lora-overhead-with-batch-size}
\end{figure}

Although LoRA greatly reduces memory usage by introducing only a small number of trainable parameters, it leads to significant runtime overhead in practice.
Figure~\ref{fig:motivation-lora-overhead-with-batch-size} compares the throughput of a frozen linear layer ($n$=$k$=4096) against its LoRA-equipped linear layer with different numbers of tokens and ranks.
The dark blue lines represent the throughput of the frozen linear layer, and other lines represent the throughput when LoRA modules are applied.
We make several key observations:
First, the throughput of LoRA linear modules is consistently lower than that of the frozen linear layer, exhibiting a slowdown of approximately 40\% and 36\% for forward and backward passes, respectively, regardless of the number of tokens.
Second, \texttt{torch.compile}~\cite{ansel2024pytorch-compile}, which provides compiler-based fusion capabilities in PyTorch, provides zero benefits in the forward pass and only negligible improvements in the backward pass.
Third, the choice of LoRA rank ($r$=16 or $r$=32) also minimally impacts throughput, indicating that the overhead is dominated by inefficient memory access patterns rather than algorithmic cost.

\begin{figure}[t]
    \centering
    \includegraphics[width=0.99\linewidth]{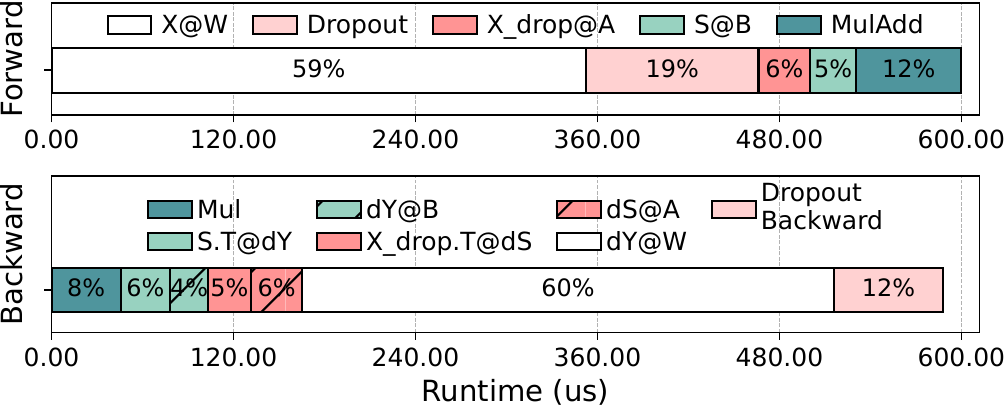}
    \caption{Runtime breakdown of a LoRA linear module with $n$=$k$=4096, $r$=16, and $tokens$=8192. \texttt{@} is matrix multiplication, \texttt{d} indicates a gradient, and \texttt{.T} represents a transpose.}
    \label{fig:motivation-lora-overhead-breakdown}
\end{figure}

The above profiling results are surprising. 
In theory, the LoRA adapter should only incur minimal FLOPs and memory accesses overhead since $r \ll n,k$.
To investigate the source of the overhead, we perform detailed profiling on both the layer-level and kernel-level.
As shown in Figure~\ref{fig:motivation-lora-overhead-breakdown}, the total runtime is dominated by three categories of operations: (1) GEMM operations of the frozen linear layer ($XW$), (2) GEMM operations from LoRA modules (down- and up-projections $\widehat{X}A$ and $SB$), and (3) other element-wise operations. 
Although the original GEMM operation still dominates runtime (59\% and 60\% for forward and backward passes, respectively), the LoRA-specific GEMM operations and element-wise computations introduce substantial overhead.
Specifically, LoRA GEMM operations account for 10.76\% and 20.37\% of overall runtime for forward and backward passes, respectively.
These kernels are memory-bandwidth-bound because of the small rank $r$, and thus the memory read and write of the large activation tensor become the bottleneck.
A simple analysis shows that the arithmetic intensity $\mathbb{I}$ of LoRA's down-projection GEMM operation ($\widehat{X}A$) in half-precision is:
\begin{equation}
        \text{Arithmetic Intensity $\mathbb{I}$} =\frac{1}{\frac{1}{r}+\frac{1}{n}+\frac{1}{m}} \ll \mathbb{B}
\end{equation}
where $r$, $n$, and $m$ denote the LoRA rank, output dimension, and batch size, respectively. This intensity $\mathbb{I}$ is far below the machine balance $\mathbb{B}$ (e.g. $\sim$295 for FP16 on NVIDIA H100 GPUs) because of the small $r$, confirming that performance is bottlenecked by memory bandwidth rather than compute throughput.
Moreover, the additional element-wise operations, including dropout, element-wise multiplication, and addition of the partial results from the branches, take 30.46\% and 17.49\% of the total execution time.
These operations are also memory-bound because of the large size of the input and output activation tensors.
To further quantify the memory impact, we profile the kernels using NVIDIA Nsight Compute~\cite{nsight-compute}, which shows that the total GPU global memory read/write traffic increases by approximately 2.64$\times$ compared to the original frozen linear layer.

Therefore, we conclude that the runtime overhead is primarily due to the redundant memory accesses of the large activation tensor relative to their small computational scale.
This analysis highlights that while the additional parameters and FLOPs introduced by LoRA seem negligible, the runtime overhead is significant due to the extra memory access.


\subsection{Overlooked Opportunities in Multi-LoRA}
\label{sec:motivation-multi-lora-oversight}

Multi-LoRA techniques refer to grouping multiple LoRA adapters sharing the same base model from separate tasks into a single batched operation. 
Such techniques have been widely adopted in LLM serving scenarios to improve GPU utilization by mitigating memory-bandwidth-bound bottlenecks of the frozen GEMM operations caused by small token counts during decoding~\cite{chen2024punica,sheng2023s-lora,wu2024dlora}.
However, multi-LoRA techniques are rarely used in fine-tuning. Although mLoRA~\cite{ye2023mlora} and Zheng et al.~\cite{zheng2024online} initially used multi-LoRA grouping to reduce the memory footprint of replicated pre-trained models and improve training efficiency, its broader implications for performance optimization remain overlooked.
In this section, we first identify two key opportunities where multi-LoRA can significantly improve training efficiency: reducing distributed training overhead and improving GPU load balance. We also explicitly analyze the limitations of existing multi-LoRA techniques.

\begin{figure}[t]
    \centering

    \begin{minipage}[c]{0.49\linewidth}
        \centering
        \includegraphics[width=\linewidth]{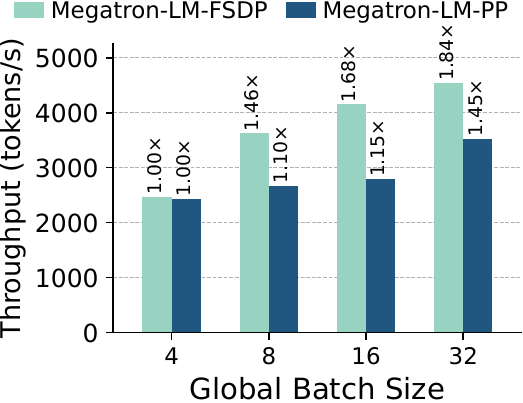}
        \caption[Ideal throughput of LLaMa-3.1-70B on 4 H100 GPUs vs. global batch sizes.]{
            Ideal throughput\footnotemark of LLaMa-3.1-70B on 4 H100 GPUs vs. global batch sizes.
        }
        \label{fig:motivation-gbz-perf}
    \end{minipage}
    \hfill
    \begin{minipage}[c]{0.49\linewidth}
        \centering
        \includegraphics[width=\linewidth]{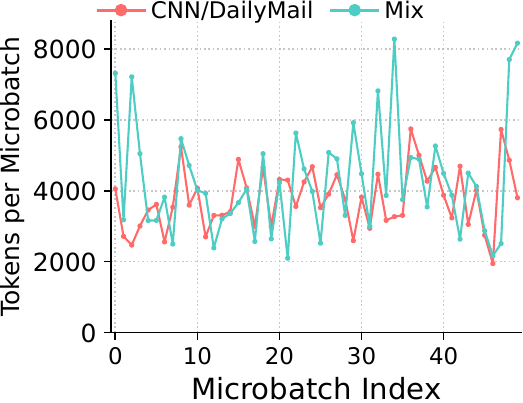}
        \caption{Number of tokens per micro-batch with a fixed micro batch size = 4.}
        \label{fig:motivation-imbalance}
    \end{minipage}
\end{figure}
\footnotetext{"Ideal" assumes uniform tokens per microbatch and no load imbalance.}

\myparagraph{Mitigating Distributed Parallelism Overhead}
By grouping adapters from multiple jobs, multi-LoRA can significantly increase global batch sizes with independent groups of tokens.
As Figure~\ref{fig:motivation-gbz-perf} illustrates, increasing global batch size from 4 to 32 enhances ideal throughput by 84\% and 45\% for FSDP and PP, respectively.
This improvement occurs because larger batch sizes improve computation-communication overlap in FSDP and reduce pipeline bubbles in PP, significantly reducing the overhead of distributed parallelism.
Additionally, since adapters from independent jobs have no interdependencies, multi-LoRA naturally enables near-zero pipeline bubbles by fully utilizing pipeline stages.
While mLoRA~\cite{ye2023mlora} primarily focused on reducing memory usage on mid-range clusters and implementing uniform adapter filling in pipeline parallelism, our approach broadens these insights to address overhead in general distributed parallelism scenarios.

\begin{figure}[t]
    \centering
    \includegraphics[width=0.99\linewidth]{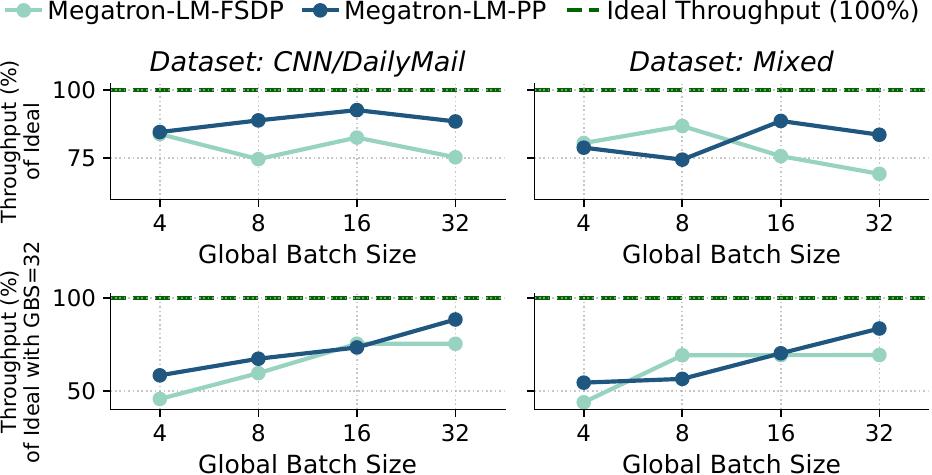}
    \caption{Performance slowdown of practical LoRA fine-tuning of LLaMa-3.1-70B on 4 H100 GPUs compared to ideal fixed-length distributed training scenarios.}
    \label{fig:motivation-practical-slowdown}
\end{figure}

\myparagraph{Reducing Load Imbalance Across GPUs}
In ideal scenarios, tokens per micro-batch are uniform, but real workloads often have significant variations, causing load imbalance. 
Figure~\ref{fig:motivation-imbalance} shows token counts per micro-batch size of 4 for two datasets: CNN/DailyMail~\cite{cnn-dailymail-dataset}, a common summarization dataset, and a Mix combining three summarization datasets (XSum~\cite{xsum-dataset}, CNN/DailyMail~\cite{cnn-dailymail-dataset}, and WikiSum~\cite{wikisum-dataset}). 
Detailed length distributions are provided in Figure~\ref{fig:evaluation-fig-0-data-statistics}. 
The substantial variation in tokens per micro-batch creates critical load imbalance in distributed training, limiting performance to the slowest GPU. In FSDP, different ranks process different microbatches but must synchronize before each layer, while in pipeline parallelism, imbalance creates pipeline bubbles across stages. 
Both scenarios degrade performance. 
As shown in the top sub-figures of Figure~\ref{fig:motivation-practical-slowdown}, practical LoRA fine-tuning experiences a significant slowdown (up to $\sim$30\%) compared to ideal fixed-length distributed training when token counts are imbalanced.

With multi-LoRA fine-tuning, the global batch size includes more samples, creating an opportunity to batch them in a way that balances token counts per microbatch. This mitigates the impact of sequence length variability. 
Figure~\ref{fig:motivation-practical-slowdown} (bottom) further illustrates the theoretical ideal throughput improvements achievable (up to 2.28$\times$) by effectively addressing both inefficiencies through multi-LoRA fine-tuning.

\myparagraph{Limitations of mLoRA}
While mLoRA~\cite{ye2023mlora} represents an important step towards multi-LoRA fine-tuning, its design has several limitations that hinder performance and scalability.
First, mLoRA assumes uniform adapter grouping and schedules according to memory capacity, but does not handle load imbalance from variable sequence lengths found in real workloads. Furthermore, its \texttt{BatchLoRA} kernel reduces kernel launch overhead but still does not solve the core memory redundant access bottleneck identified in Section~\ref{sec:motivation-lora-overhead}.
Finally, mLoRA's design is narrowly focused on Pipeline Parallelism and relies on inefficient CPU-based communication, which limits scalability on modern GPU clusters with high-bandwidth interconnects like NVLink~\cite{nvlink}. 

\section{Overview and Key Ideas}
\label{sec:overview-and-key-ideas}

To address the two key inefficiencies identified in Section~\ref{sec:motivation}: runtime overhead from redundant memory access and throughput degradation due to load imbalance and parallelism overhead, we propose \textit{\Name}, a novel LLM LoRA fine-tuning system that improves system throughput via kernel-level optimizations and job-level scheduling.
In this section, we summarize the two key ideas behind it:

\begin{figure}[t]
    \centering
    \includegraphics[width=0.975\linewidth]{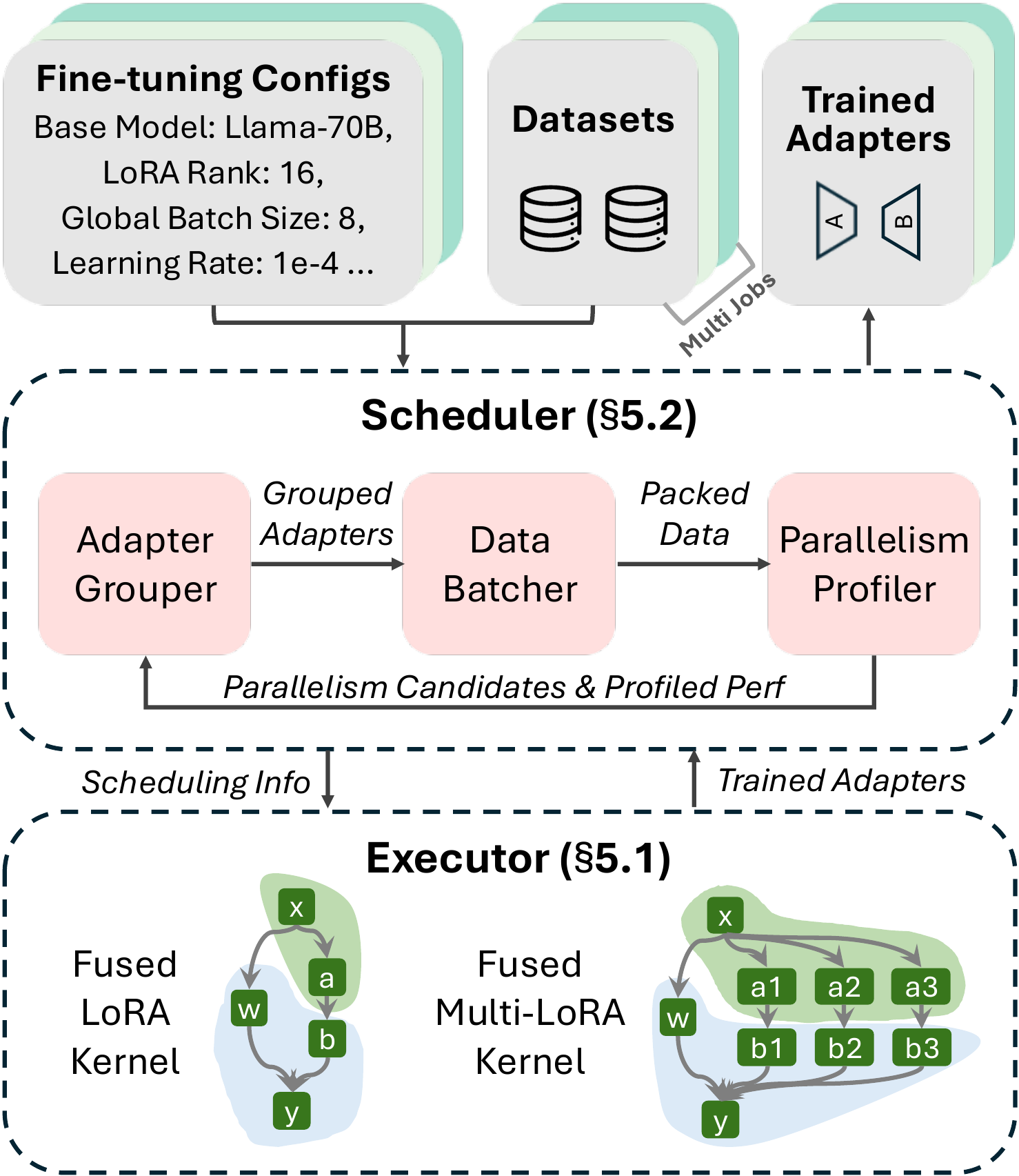}
    \caption{Overview of LoRAFusion.}
    \label{fig:key-ideas-overview}
\end{figure}

\myparagraph{1. FusedLoRA and FusedMultiLoRA}
To reduce LoRA's runtime overhead, \Name fuses memory-bound operations while preserving performance for compute-bound matrix multiplications by carefully splitting the computation graph.
This is motivated by the key bottleneck in LoRA, i.e., redundant memory accesses to large activation tensors. 
A naive solution is to fuse the entire computation graph into a single kernel. 
However, doing so introduces costly recomputation or synchronization overhead~\cite{shi2023welder}. It also consumes GPU resources like registers and shared memory, which degrades GEMM performance due to suboptimal tiling.

\Name addresses this by introducing a graph-splitting strategy. 
Instead of full fusion, it splits the graph at the intermediate tensors with LoRA rank $r$, which are small and cheap to materialize. 
This enables fusion around full-sized activations without recomputation or synchronization.
Thus, the resulting FusedLoRA kernels reduce memory traffic while preserving the performance of frozen GEMMs.

We further extend this idea to support multiple LoRA adapters using a tile-level routing mechanism.
The proposed FusedMultiLoRA processes tokens from different jobs within the same fused kernel.
Each tile uses a precomputed mapping to dynamically select the appropriate adapter weights.
This avoids the need for separate kernel launches per adapter and maintains high GPU utilization across jobs.
During backpropagation, gradients are routed similarly. The FusedMultiLoRA forms the foundation of \Name's job-level scheduling.

\myparagraph{2. Multi-LoRA Scheduling}
To exploit the benefits of multi-LoRA fine-tuning, \Name introduces a scheduler that coordinates adapter grouping and data batching.
The major challenge is to construct well-balanced microbatches across jobs while respecting data dependencies between global batches. Specifically, in pipeline parallelism, each global batch must wait for all samples from the previous one to complete their backward passes.
\Name addresses this with a two-stage hierarchical strategy: it first groups adapters in a way that keeps their batches spread apart in the schedule, then solves a bin-packing problem to batch their samples and reduce imbalance.

In the first stage, \Name groups LoRA adapters based on their sample length distributions to ensure that consecutive global batches from the same adapter are sufficiently spaced in the schedule.
In the second stage, samples within each group are packed into microbatches using a two-step MILP-based optimization, with the greedy algorithm as a fallback and multiprocessing for efficiency.
A final merge pass reduces underfilled microbatches when possible.
This scheduling strategy improves load balance and reduces pipeline stalls, boosting overall system throughput.


\myparagraph{System Workflow}
Figure~\ref{fig:key-ideas-overview} presents \Name's system workflow. Given a set of fine-tuning jobs, \Name first extracts dataset statistics and proposes a microbatch token budget via a parallelism simulator. It then forms adapter groups and constructs microbatches accordingly. The grouping and batching outputs are re-evaluated through simulation, and the process iterates until a high-throughput configuration is found. Finally, jobs are executed using the fused kernels described above. A multi-adapter runtime coordinator ensures token-to-adapter consistency, manages resource sharing, and tracks gradients across job boundaries.
Through the combination of fused execution and coordinated job scheduling, \Name addresses both the memory and parallelism bottlenecks in LoRA fine-tuning, improving throughput while maintaining correctness and generality.

\section{System Design}
\label{sec:system-design}
\subsection{FusedLoRA and FusedMultiLoRA}
\label{sec:fused-lora-and-fused-multi-lora}

As described in Section~\ref{sec:motivation-lora-overhead}, LoRA modules introduce significant runtime overhead despite adding only a small number of parameters. 
Our profiling reveals that this overhead stems primarily from redundant memory access. 
Our goal is to fuse operations within the LoRA computation graph to reduce these memory transfers while preserving compute efficiency.

\myparagraph{Design Considerations and Challenges}
\begin{sloppypar}
While kernel fusion can reduce redundant memory access, fusing all LoRA operations into a single kernel introduces practical challenges.
First, the frozen GEMM operation $Y_1 = XW$ is compute-bound and highly sensitive to kernel tiling strategies and GPU resource usage (e.g., shared memory and register file). 
A suboptimal tiling layout or overuse of registers and shared memory can greatly degrade its performance.
Second, fusing operations with producer-consumer dependencies, such as $\widehat{X}A$ followed by $(\widehat{X}A)B$, may require recomputing intermediate results or introducing thread block synchronization, both of which can add overhead~\cite{shi2023welder}.
Thus, a good fusion strategy must minimize memory access while preserving optimal compute performance and avoiding expensive synchronization or recomputation.
\end{sloppypar}

\begin{figure}[t]
    \centering
    \includegraphics[width=0.99\linewidth]{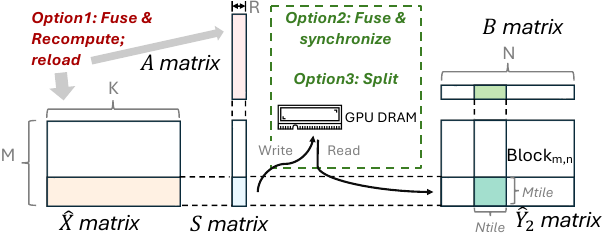}
    \caption{Overview of our fusion strategy for LoRA modules in the forward pass, illustrating the full graph fusion approach vs. the split graph fusion approach.}
    \Description{Fusion design.}
    \label{fig:system-design-fig-2-fusion-design}
\end{figure}

\myparagraph{Full Graph Fusion vs. Split Graph Fusion} 
Figure~\ref{fig:system-design-fig-2-fusion-design} illustrates three design choices for handling the intermediate tensor $S = \hat{X}A$ in the forward pass. 
The first option recomputes $S$ inside each tile of the fused kernel, but requires loading the entire $A$ matrix repeatedly, becoming expensive when batch size $M$ is large. 
The second option fuses computation and uses synchronization across thread blocks to share $S$, where only a single $Mtile$ (the m-th block in the token dimension) computes the intermediate $S$ tiles and writes to global memory, while other tiles wait through a semaphore. This adds coordination overhead. 
Our approach takes a third option: explicitly storing and reloading $S$ from GPU global memory. Since $S$ is much smaller than other tensors and depends on the small LoRA rank $r$, the cost of reading and writing it is low. 
Splitting the graph at $S$ avoids both recomputation and synchronization while reducing expensive memory traffic associated with full-sized activation tensors. 
This approach preserves GPU resources for optimal tiling of the compute-bound $XW$ operation, maintaining peak performance for the most computationally intensive part.

\begin{figure*}[t]
    \centering
    \includegraphics[width=0.99\linewidth]{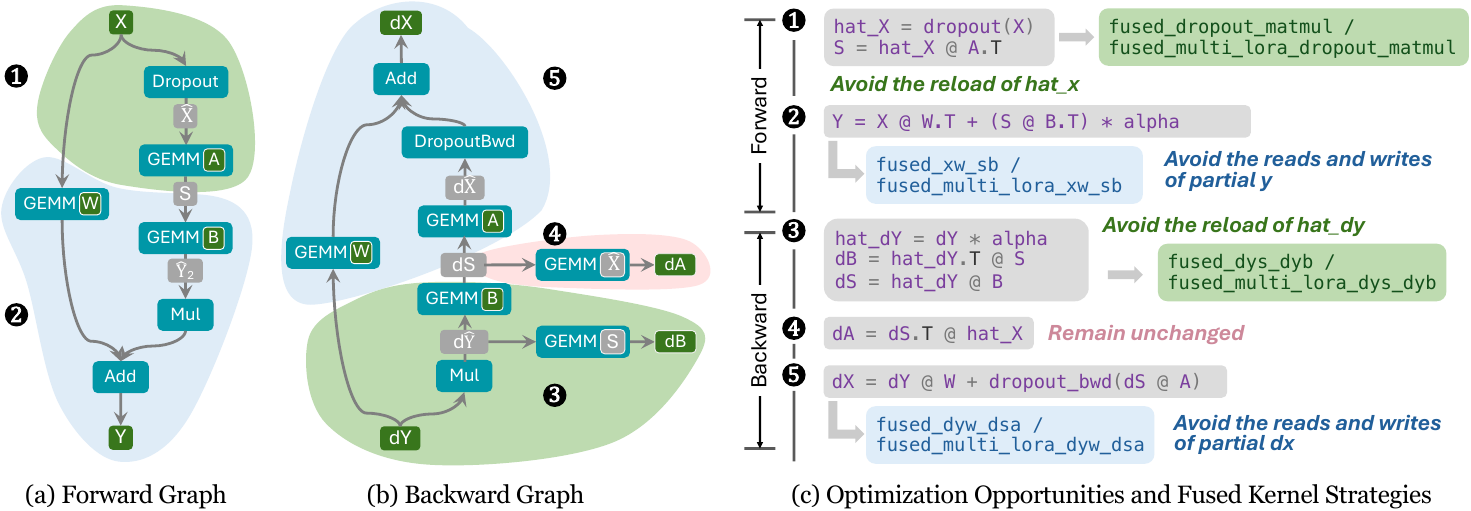}
    \caption{LoRA kernel design. FusedLoRA reduces memory accesses by combining memory-heavy LoRA branches with base GEMM operations on frozen weights. The right figure has transposed weight tensors to match the hardware memory layout.}
    \label{fig:system-design-fig-1-lora-kernel}
\end{figure*}

\myparagraph{FusedLoRA Design}
Figure~\ref{fig:system-design-fig-1-lora-kernel} illustrates our fused kernel design for both forward and backward passes.
In the forward pass (Figure~\ref{fig:system-design-fig-1-lora-kernel}(a)), 
we combine dropout and down-projection into a single kernel (\filledone) to eliminate reloading of the full-sized activation tensor.
We also fuse the compute-bound base model GEMM ($Y_1 = XW$) with the memory-bound LoRA operations ($Y_2 = \alpha SB$) (\filledtwo). 
This fusion eliminates redundant memory operations and saves one read and write of the full-sized output tensor by directly accumulating partial results, without affecting the base GEMM performance.
In the backward pass, we apply similar principles. Operation \filledthree fuses the gradient computation of $S$ and $B$, eliminating the need to reload $dY$.
Operation \filledfour remains separate since it operates on the small masked input, where fusion provides minimal benefit.
Operation \filledfive horizontally fuses the compute-intensive gradient computation for the base model with memory-bound LoRA path operations, preventing redundant reads and writes of partial output gradients.
The key insight of our approach is strategically identifying operations where horizontal fusion reduces memory traffic without compromising computational efficiency. By fusing operations that share large tensors (\filledtwo, \filledthree, and \filledfive), we significantly reduce memory bottlenecks while maintaining optimal tiling strategies for compute-bound operations.

\begin{figure}[t]
    \centering
    \includegraphics[width=0.975\linewidth]{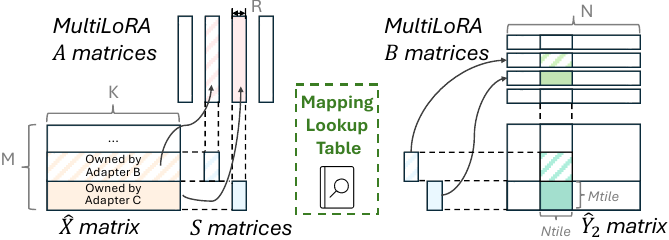}
    \caption{Illustration of FusedMultiLoRA in the forward pass. The routing of LoRA adapters is done at the tile level.}
    \Description{FusedMultiLoRA.}
    \label{fig:system-design-fig-3-fused-multi-lora}
\end{figure}

\myparagraph{Extending to FusedMultiLoRA}
To support concurrent fine-tuning of multiple LoRA adapters, we extend FusedLoRA to FusedMultiLoRA, allowing our fused kernels to operate on mixed-adapter batches from different jobs. 
As shown in Figure~\ref{fig:system-design-fig-3-fused-multi-lora}, each input $Mtile$ is tagged with an adapter ID and configuration, such as LoRA rank, scaling factor, and dropout ratio, stored in a lightweight lookup table. 
During execution, the frozen model computation is shared across all tokens, while adapter-specific logic is applied dynamically per $Mtile$. 
For each ($Mtile, Ntile$) tile of the output, the kernel loads the appropriate $A$ and $B$ matrices and applies the correct scaling and dropout. 
In the backward pass, the same mapping is used to route gradients to their respective adapters without interference. 
This tile-level routing allows efficient execution of heterogeneous adapters in a single fused run, avoiding redundant computation and enabling the job-level optimizations introduced in Section~\ref{sec:scheduler}.

FusedLoRA reduces memory traffic by fusing LoRA operations around shared activations while preserving base model efficiency. Building on FusedLoRA, FusedMultiLoRA supports heterogeneous adapters via tile-level routing. The system dynamically chooses between them, falling back to FusedLoRA when only one adapter is present in the batch.
\subsection{Multi-LoRA Scheduler}
\label{sec:scheduler}

\begin{figure}[t]
    \centering
    \includegraphics[width=0.93\linewidth]{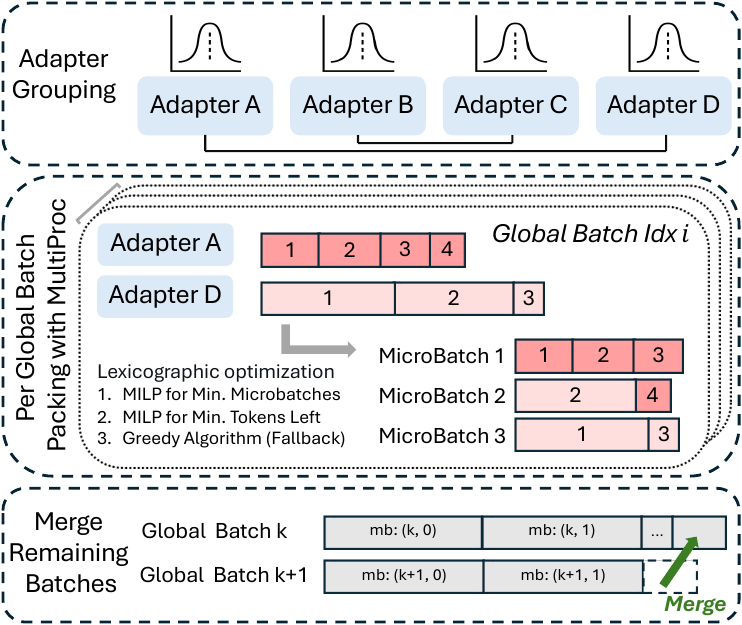}
    \caption{Multi-LoRA adapter scheduling workflow. Top: Adapter grouping by sequence length statistics. Middle: Two-stage MILP optimization for microbatch creation. Bottom: Cross-batch merging of underfilled microbatches.}
    \label{fig:system-design-fig-4-scheduler}
\end{figure}

\Name not only reduces runtime overhead through fused kernels but also improves end-to-end throughput by scheduling multiple LoRA fine-tuning jobs together.
This is achieved by grouping adapters and adaptively batching their samples to balance GPU load and minimize distributed parallelism overhead.
Figure~\ref{fig:system-design-fig-4-scheduler} shows the overall process. 
Adapters are grouped based on sequence length statistics (top), and adapters in the same group are trained jointly.
For each group, we aggregate samples into global batches and pack each one into microbatches using a two-stage MILP-based optimization (middle).
Once microbatches are generated in parallel with multiprocessing, a final merge pass combines underfilled microbatches across global batches when data dependencies allow (bottom).

\myparagraph{Granularity}
Due to data dependencies between consecutive global batches, our scheduling operates at the granularity of individual global batches. Each adapter's dataset is divided into global batches based on the user-specified global batch size. We then aggregate all samples belonging to the same global batch index across adapters and pack them into multiple microbatches.

\myparagraph{Bubble Lemma \& Adapter Grouping}
\Name first groups LoRA adapters before batching samples to reduce scheduling complexity.
In pipeline parallelism with $S$ stages, a sample's backward pass begins only after $S - 1$ other microbatches complete forward passes.
To maintain data dependencies, we define the \textit{bubble lemma}:
for adapter $i$, if sample $s$ from global batch $j$ is committed at microbatch $k$, no sample from batch $j$+1 of the same adapter can be committed before microbatch $k$+$S$-1 (after sample $s$'s backward pass completes).
Without constraints, adaptive batching might scatter samples from consecutive batches across microbatches.
The tail of batch $j$ could conflict with batch $j$+1's head, causing incorrect execution.
We resolve this conflict by grouping adapters with strict ordering between groups, while allowing flexible merging within each group.
This ensures batches from the same adapter are separated by microbatches from other groups, satisfying the bubble condition.
For load balance within groups, we use head-tail pairing, sorting adapters by mean token length, and pairing short-sequence adapters with long ones.
This grouping balances constraints with flexibility and forms our data packing foundation.

\begin{algorithm}[t]
    \footnotesize
    \caption{Data Batching \& Merging (Per Group)}
    \label{alg:multi-lora-data-batching}
    \DontPrintSemicolon
    \KwData{Adapters with grouped samples, token capacity $C$, timeout $t$}
    \KwResult{Scheduled microbatches satisfying pipeline constraints}
    
    \ForEach{global batch $b$ in parallel}{
        \tcp{Greedy fallback as baseline}
        $(B_g, \{m_i^g\}) \gets \texttt{GreedyPacking}(b, C)$
        
        \tcp{Stage 1: minimize number of microbatches}
        $B^* \gets \texttt{MILP\_MinBins}(b, C, \text{timeout}=t)$ \;
        \If{$B^* \geq B_g$}{
            $B^* \gets B_g$
            }
            
            \tcp{Stage 2: minimize smallest bin tokens}
            $\{m_i\}_{i=1}^{B^*} \gets \texttt{MILP\_MinSmallestBin}(b, B^*, C, \text{timeout}=t)$ \;
            \If{$B^* = B_g$ \textbf{and} $\min_i m_i \geq \min_i m_i^g$}{
                \Return $\texttt{GreedyPacking}(b, C)$
                }
                }
                
                \ForEach{consecutive batch pairs $(b, b{+}1)$}{
                    Shift tokens from $b{+}1$ into $b$ if bubble lemma is preserved
                    }
                    
                    \texttt{VerifyAndFix(schedule)} \tcp*{Insert no-ops where needed}
                    
    \KwRet Scheduled microbatches
\end{algorithm}

\myparagraph{Data Batching with Two-Stage MILP}
After adapters are grouped, we solve a bin-packing problem to batch samples into microbatches, each constrained by a fixed token capacity.
Our goal is to reduce both the total number of microbatches and the impact of underfilled ones, which affect load balance and pipeline utilization. Specifically, we aim to (i) minimize the number of microbatches needed to pack all samples, and (ii) make the smallest microbatch as empty as possible to enable better merging in later stages.

We address the bin-packing problem using a two-stage mixed-integer linear programming (MILP) formulation (see Algorithm~\ref{alg:multi-lora-data-batching}, lines 3-7).
For notational brevity, let $P$ denote the padding multiple, which is a user-specified parameter to pad the sequence length belonging to the same adapter to a multiple of $P$ (e.g., 64 or 128). Let $x_{s,b} \in \{0,1\}$ denote whether sample $s$ is assigned to bin $b$, $k_{a,b} \in \mathbb{N}$ be the padded multiples contributed by adapter $a$ in bin $b$, and $z_b \in \{0,1\}$ indicate whether bin $b$ is used.  In the first stage, we solve:
\begin{equation}
\begin{aligned}
    \argmin_{x_{s,b}, k_{a,b}, z_b} & \quad \sum_{b=1}^B z_b \\
    \text{s.t.} \quad & \quad z_{b+1} \leq z_b \quad \forall b < B \\
    &\quad \sum_{b=1}^B x_{s,b} = 1 \quad \forall s \in \text{samples} \\
    &\quad \sum_{s \in \text{adapter}(a)} \text{len}(s) \cdot x_{s,b} \leq k_{a,b} \cdot P \quad \forall a,b \\
    &\quad  z_b \leq \sum_{a} k_{a,b} \cdot P \leq \text{capacity} \cdot z_b \quad \forall b
\end{aligned}
\end{equation}
The constraints ensure used bins are contiguous from the start, each sample is assigned to exactly one bin, adapter-specific token counts respect padding multiples, and bin capacity is not exceeded.

With the optimal number of bins $B^{*}$ from the first stage, the second stage fixes $B = B^{*}$ and minimizes the smallest total token count among all bins. The second stage solves:
\begin{equation}
\begin{aligned}
    \argmin_{x_{s,b}, k_{a,b}} & \quad \min_{b \in [1, B^*]} \sum_{a} k_{a,b} \cdot P \\
    \text{s.t.} \quad & \quad \sum_{b=1}^{B^*} x_{s,b} = 1 \quad \forall s \in \text{samples} \\
    &\quad \sum_{s \in \text{adapter}(a)} \text{len}(s) \cdot x_{s,b} \leq k_{a,b} \cdot P \quad \forall a,b \\
    &\quad \sum_{a} k_{a,b} \cdot P \leq \text{capacity} \quad \forall b \\
\end{aligned}
\end{equation}
This optimization problem can be reformulated into an MILP problem. It minimizes the smallest bin size, which leaves more space in the least-full microbatch for potential merging in later stages, thereby reducing pipeline stalls. 

To improve runtime efficiency, we implement two techniques (also reflected in Algorithm~\ref{alg:multi-lora-data-batching}). First, we set a timeout on the MILP solver and fall back to a greedy bin-packing algorithm if the solver takes too long (lines 2, 5, and 9). Second, since global batches are independent, we parallelize the bin-packing optimization across batches using multiprocessing (line 1), which allows us to efficiently schedule all training data while balancing load and reducing microbatch count.

\myparagraph{Merging \& Verification}
After microbatch packing, the final microbatch in a global batch may be underfilled, reducing GPU efficiency and increasing pipeline bubbles. To mitigate this issue, we apply a greedy merge pass that shifts tokens from the next global batch into the current batch's final microbatch (as shown in Figure~\ref{fig:system-design-fig-4-scheduler} bottom), as long as capacity and the bubble lemma are satisfied (Algorithm~\ref{alg:multi-lora-data-batching}, lines 12-14). We then perform a verification step to ensure no constraint is violated. If any bubble condition is unmet, we insert no-op microbatches to restore correctness and preserve pipeline consistency (line 15).

\myparagraph{Parallelism Profiler}
The scheduler requires token capacity as input, which depends on the parallelism strategy.
\Name assumes that effective scheduling keeps tokens within each microbatch close to the token capacity.
Since token capacity and parallelism strategies are orthogonal to scheduling, they should be tuned outside the scheduler using automatic parallelization techniques~\cite{zheng2022alpa,flexflow,liu2024aceso,zhu2025mist}.
We implement a lightweight profiler that directly benchmarks runtime under different model parallelism configurations with fixed-length inputs and collects throughput. 
We choose the best-performing configuration and pass its token capacity to the data batching stage, ensuring that microbatch packing aligns with the system's performance characteristics.

\section{Evaluation}
\label{sec:evaluation}

\sloppy

We implement \Name with $\sim$10K LoC in Python. 
The FusedLoRA and FusedMultiLoRA kernels are developed using Triton~\cite{tillet2019triton}, and multi-adapter pipeline parallelism is built on top of Megatron-LM~\cite{megatron-1}.
Since Megatron-LM does not natively support LoRA, we integrate Hugging Face Transformers~\cite{hf-transformers} and the PEFT Library~\cite{peft} for model architecture and LoRA adaptation.

The optimizations in \Name are designed to be lossless, guaranteeing they do not affect model convergence or final quality.
Our FusedLoRA and FusedMultiLoRA kernels are numerically stable, producing outputs that are functionally identical to the baseline implementations within numerical precision.
While our adaptive scheduler rearranges samples to form balanced microbatches, it strictly preserves the order of global batches, ensuring the sequence of gradient updates remains unchanged. 
Given that model behavior is identical by design, our evaluation focuses exclusively on system performance metrics like throughput.

We evaluate \Name across a range of real-world fine-tuning scenarios involving multiple datasets, model scales, GPU platforms, and job configurations.
Our primary metric is throughput, measured in trained tokens per second, which better reflects system efficiency for inputs with sequence length variations.
Finally, we perform detailed scalability studies, ablation studies, and performance breakdowns to analyze the contribution of each system component.

\begin{figure}[t]
    \centering
    \includegraphics[width=0.975\linewidth]{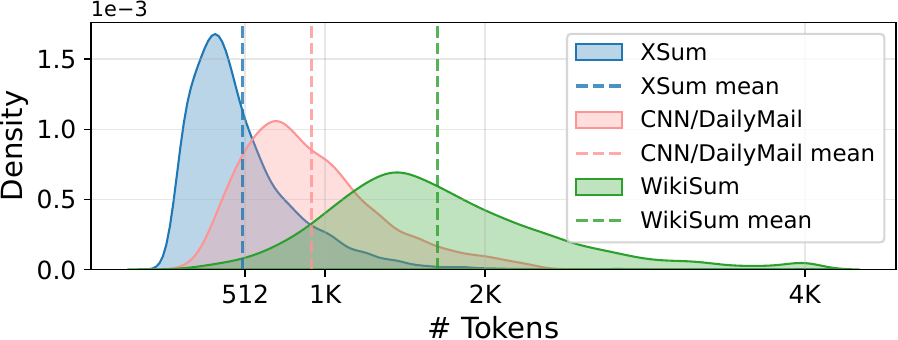}
    \caption{Distribution of sample lengths across the XSum~\cite{xsum-dataset}, CNN/DailyMail~\cite{cnn-dailymail-dataset}, and WikiSum~\cite{wikisum-dataset} datasets used for LoRA fine-tuning.}
    \label{fig:evaluation-fig-0-data-statistics}
\end{figure}

\subsection{Methodology}
\label{sec:evaluation-methodology}

\myparagraph{Hardware Settings} 
We primarily benchmark \Name on a GPU cluster with NVIDIA H100 (80GB) GPUs and additionally report results on L40S (48GB) GPUs to demonstrate generalizability.
Each H100 node is equipped with 8$\times$H100 GPUs connected via NVLink, 208 vCPUs, and Infiniband for multi-node communication. 
Each L40S server contains 4$\times$L40S GPUs connected over PCIe and 128 vCPUs. 
Most experiments use the smallest number of GPUs that fit the model and maintain good utilization, typically 1, 2, or 4 GPUs. 
As shown in our Scalability Studies (Section~\ref{sec:evaluation-scalability-studies}), assigning fewer GPUs per job and using additional GPUs to run more independent jobs often leads to better efficiency, as it reduces inter-GPU communication and synchronization overhead.
The Scalability Studies also show \Name is fully compatible with both data-parallel and multi-node scaling.

\myparagraph{Workload Settings} 
We evaluate \Name on three open-source language models of varying sizes: LLaMa-3.1-8B~\cite{llama3.1}, Qwen-2.5-32B~\cite{yang2024qwen2-5}, and LLaMa-3.1-70B~\cite{llama3.1}. 
All experiments use summarization as a representative sequence-to-sequence task, which is widely used in prior LoRA fine-tuning studies~\cite{hu2022lora, li2023loftq, liu2024rst}.
We select three public summarization datasets: XSum~\cite{xsum-dataset}, CNN/DailyMail~\cite{cnn-dailymail-dataset}, and WikiSum~\cite{wikisum-dataset}. These datasets have diverse length distributions, as shown in Figure~\ref{fig:evaluation-fig-0-data-statistics}, which stresses batching and scheduling under realistic conditions.
For multi-LoRA experiments, we train four LoRA adapters in parallel. 
In the XSum, CNN/DailyMail (CNNDM), and WikiSum configurations, all four adapters are trained independently on the same dataset. In the Mixed setting, each adapter is trained on a dataset combining samples from all three. In the Heterogeneous (Het) setting, the four adapters are trained on different datasets: one each on XSum, CNN/DailyMail, WikiSum, and Mixed.

\myparagraph{Baselines} 
We compare \Name against three baselines: (i) Megatron-LM~\cite{megatron-1} with fully sharded data parallel (FSDP), (ii) Megatron-LM with pipeline parallelism (PP), and (iii) mLoRA~\cite{ye2023mlora}. 
Megatron-LM does not support multi-LoRA fine-tuning natively, so tasks are trained sequentially, while mLoRA supports multi-LoRA fine-tuning.
The original mLoRA uses Python RPC for inter-GPU communication, which performs poorly on NVLink-equipped GPUs. 
Therefore, we reimplement mLoRA inside our system with high-performance communication primitives to ensure fair comparison.
In addition, since mLoRA does not provide a unique multi-LoRA CUDA kernel, we optimistically assume it has the same performance as the naive single LoRA kernel.
Tensor parallelism is not evaluated due to the lack of efficient support in existing LoRA frameworks. 
All experiments use PyTorch 2.6, CUDA Toolkit 12.4, Triton 3.2.0, and Megatron-Core 0.11.0.

\subsection{End-to-End Results}
\label{sec:evaluation-end-to-end-results}

\begin{figure*}[t]
    \centering
    \includegraphics[width=0.99\textwidth]{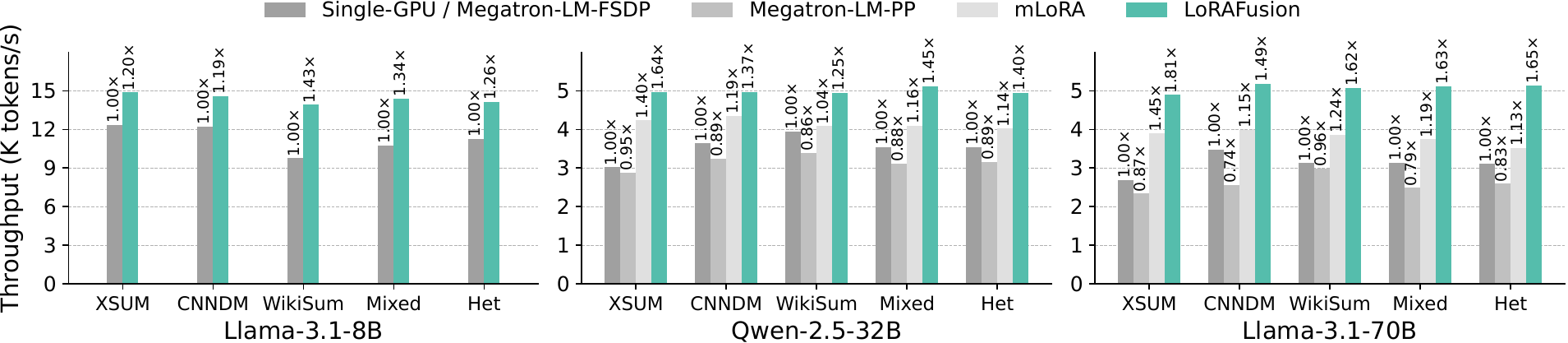}
    \caption{End-to-end training throughput (tokens/sec) of training 4 LoRA adapters on 1, 2, and 4 H100 GPUs. The first four bars per subfigure represent homogeneous workloads (same dataset), and the final (Het) shows heterogeneous adapters trained on different datasets.}
    \label{fig:evaluation-fig-1-end-to-end}
\end{figure*}

\begin{figure}[t]
    \centering
    \includegraphics[width=0.975\linewidth]{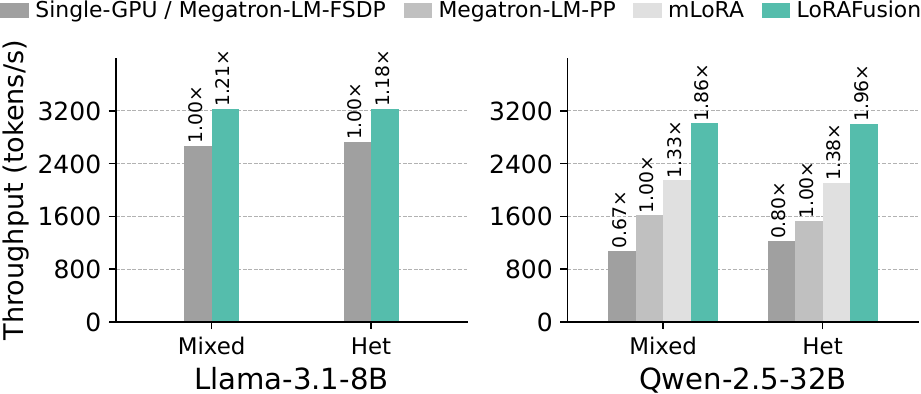}
    \caption{End-to-end training throughput (tokens/sec) of training 4 LoRA adapters on 1 and 4 L40S GPUs.}
    \label{fig:evaluation-fig-2-l40-gpus}
\end{figure}

\myparagraph{Speedup on H100 GPUs} 
Figure~\ref{fig:evaluation-fig-1-end-to-end} reports the end-to-end throughput of training 4 LoRA adapters across three models. \Name consistently outperforms all baselines by $1.19-1.96\times$.
For LLaMa-3.1-8B, which fits on a single H100 GPU, \Name achieves an average 1.26$\times$ speedup (up to 1.43$\times$), primarily from the FusedLoRA kernel, which reduces memory traffic. 
Since single-GPU setups do not suffer from load imbalance, the improvement here directly reflects kernel-level gains. \Name achieves high speedup on the WikiSum dataset due to the large variance in sample lengths. While the baseline methods suffer from out-of-memory errors, \Name achieves stable packing.
For Qwen-2.5-32B and LLaMa-3.1-70B, which require distributed training, \Name achieves 1.42$\times$ and 1.64$\times$ average speedup (up to 1.64$\times$ and 1.81$\times$) respectively. Larger models benefit more from improved scheduling, as pipeline stalls and load imbalance become more pronounced at higher parallelism. In the most challenging heterogeneous setting (Het), where each adapter uses a different dataset, \Name still achieves strong performance, highlighting its robustness.

\begin{sloppypar}
\myparagraph{Speedup on L40S GPUs} 
Figure~\ref{fig:evaluation-fig-2-l40-gpus} presents results on NVIDIA L40S GPUs. \Name achieves  $1.19-1.91\times$ average speedup for LLaMa-3.1-8B and Qwen-2.5-32B respectively. The benefit is smaller for LLaMa-3.1-8B due to limited memory capacity on a single L40S GPU, which constrains batch size and limits kernel fusion effectiveness. However, even under such constraints, \Name maintains consistent improvements, demonstrating generalizability across model sizes and hardware platforms.
\end{sloppypar}

\begin{figure}[t]
    \centering
    \includegraphics[width=0.976\linewidth]{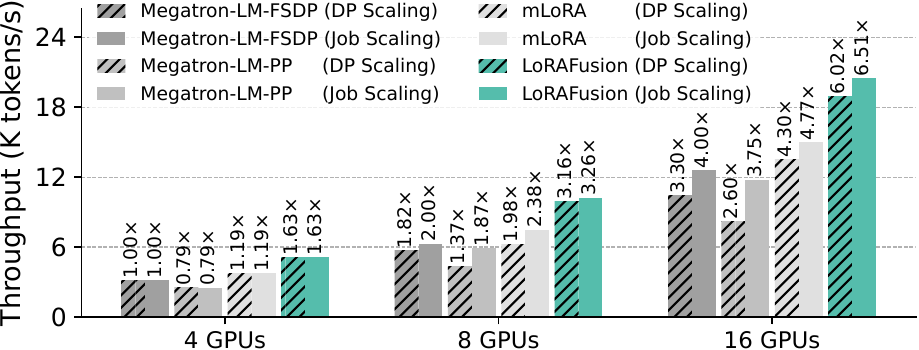}
    \caption{Scalability of \Name across 4, 8, and 16 H100 GPUs when training 4 LoRA adapters simultaneously. DP scaling means the more GPUs are used to increase the DP degree for the same job, while Job scaling means different LoRA fine-tuning jobs are scheduled to utilize more GPUs. Global batch sizes are scaled proportionally with GPU count to ensure fair comparison.}
    \label{fig:evaluation-fig-3-scaling}
\end{figure}

\subsection{Scalability Studies}
\label{sec:evaluation-scalability-studies}
We evaluate \Name on 4, 8, and 16 H100 GPUs under two scaling strategies: DP scaling (more GPUs per job) and job-level scaling (more concurrent jobs). 16 H100 GPUs experiment is conducted on 2 nodes each with 8 GPUs, connected via InfiniBand~\cite{infiniband}. Global batch size is scaled with GPU count in both settings.
We draw two key conclusions.
First, job-level scaling consistently outperforms DP scaling due to better load balance, achieving 1.18$\times$ and 1.25$\times$ higher throughput on 8 and 16 GPUs, respectively. 
Second, \Name is fully compatible with DP scaling and multi-node fine-tuning, and still delivers strong performance, achieving 1.78$\times$ average speedup over Megatron-LM and 1.50$\times$ over mLoRA under DP scaling.

\subsection{Effectiveness of FusedLoRA Kernel}
\label{sec:evaluation-effectiveness-of-fused-lora-kernel}

\begin{figure}[t]
    \centering
    \includegraphics[width=1.00\linewidth]{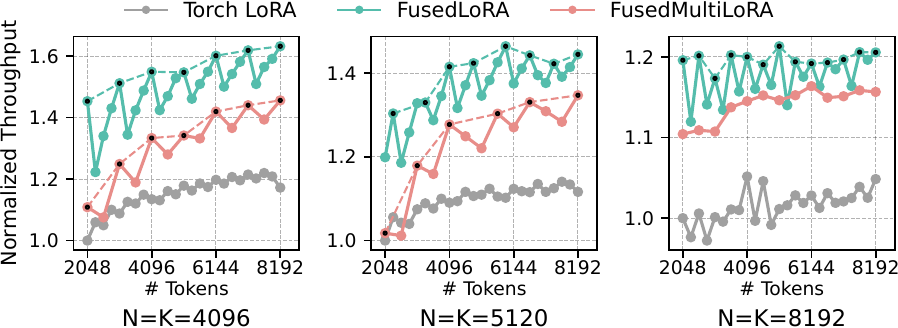}
    \caption{Performance of FusedLoRA kernel in forward and backward passes.}
    \label{fig:evaluation-fig-5-kernel-perf-fwd_bwd}
\end{figure}

\begin{figure}[t]
    \centering
    \includegraphics[width=1.00\linewidth]{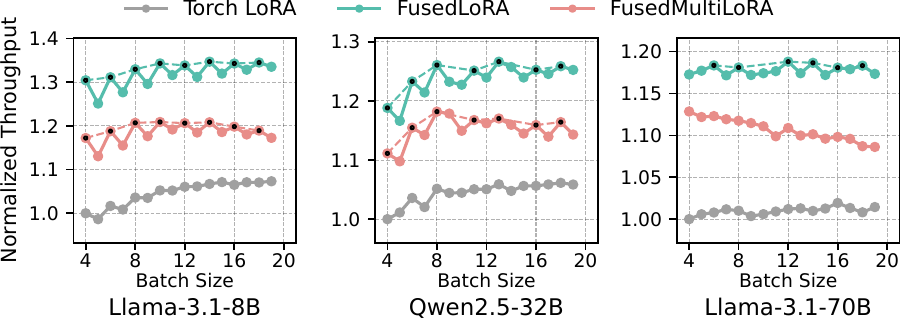}
    \caption{Performance of FusedLoRA kernel in decoder layers of different models.}
    \label{fig:evaluation-fig-6-layer-perf-normalized}
\end{figure}

\myparagraph{Kernel Performance} 
Figure~\ref{fig:evaluation-fig-5-kernel-perf-fwd_bwd} shows the throughput of our FusedLoRA and FusedMultiLoRA kernels compared to the standard Torch LoRA implementation~\cite{peft}. FusedLoRA achieves an average speedup of $1.27\times$ (up to $1.39\times$), while FusedMultiLoRA achieves $1.17\times$ on average (up to $1.24\times$). In the forward pass, FusedMultiLoRA performs similarly to FusedLoRA, as most computation is shared. In the backward pass, it incurs slight overhead from accumulating gradients across adapters and additional element-wise operations. Despite this overhead, both kernels consistently outperform the baseline across different token sizes and model configurations.

\myparagraph{Layer-wise Performance} 
Figure~\ref{fig:evaluation-fig-6-layer-perf-normalized} compares the speedup across different linear layers in the model. 
FusedLoRA achieves an average speedup of $1.21\times$ (up to $1.30\times$), while FusedMultiLoRA achieves $1.13\times$ (up to $1.17\times$). 
These results are based on microbatches containing four adapters. In practical fine-tuning workloads, each microbatch typically contains only one or two adapters, making FusedMultiLoRA's performance close to FusedLoRA.

\begin{figure}[t]
    \centering
    \includegraphics[width=0.99\linewidth]{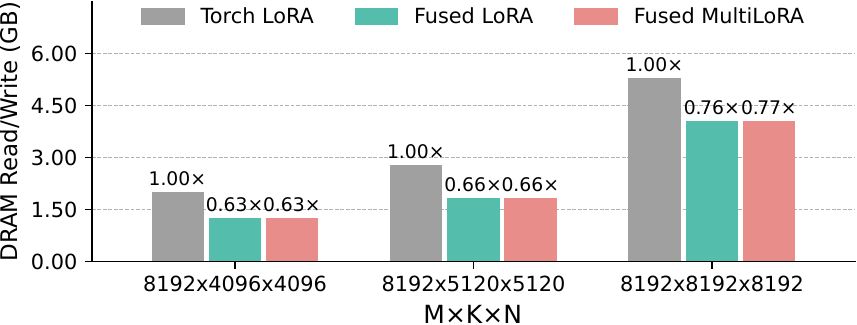}
    \caption{GPU DRAM memory traffic comparison between different kernels from NVIDIA Nsight Compute (NCU).}
    \label{fig:evaluation-fig-7-kernel-ncu-profile}
\end{figure}

\myparagraph{Memory Traffic Reduction}
Figure~\ref{fig:evaluation-fig-7-kernel-ncu-profile} shows DRAM read and write traffic from NVIDIA Nsight Compute (NCU) across representative GEMM shapes. Both FusedLoRA and FusedMultiLoRA consistently reduce memory usage compared to Torch LoRA. For example, on the $8192 \times 4096 \times 4096$ shape, total DRAM traffic reduces to $0.63\times$. Across all settings, traffic is reduced by $34\%-37\%$, confirming that our fusion design effectively reduces redundant memory access.

\myparagraph{Performance Insights Across Diverse Hardware}
The FusedLoRA and FusedMultiLoRA kernels reduce redundant memory access for large activation tensors, which is especially important on hardware where memory bandwidth is much lower compared to compute FLOPS. 
As modern accelerators increase compute FLOPS faster than memory bandwidth~\cite{gholami2024ai-and-memory-wall}, the benefits of our fused kernels are expected to grow in future systems.

\subsection{Effectiveness of Job-Level Scheduling}
\label{sec:evaluation-effectiveness-of-job-level-scheduling}

\begin{figure}[t]
    \centering
    \includegraphics[width=0.99\linewidth]{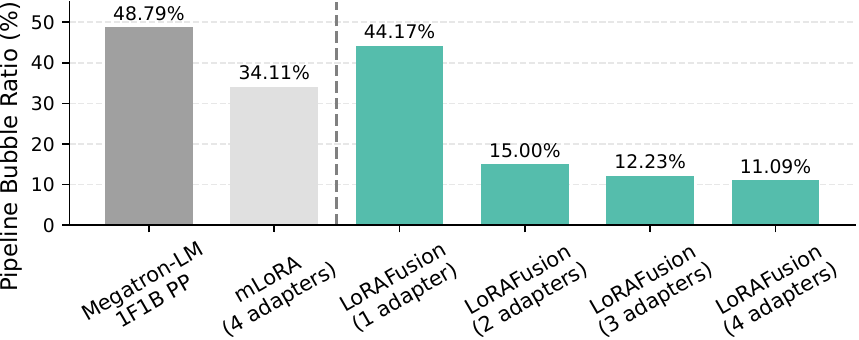}
    \caption{Pipeline bubble ratio under different methods.}
    \label{fig:evaluation-fig-8-pipeline-bubbles}
\end{figure}

\myparagraph{Pipeline Bubble Reduction} 
Figure~\ref{fig:evaluation-fig-8-pipeline-bubbles} shows how \Name helps reduce pipeline bubbles by scheduling multiple adapters together.
We make three key observations.
First, with only one adapter, the bubble ratio remains high at 44.17\%, close to Megatron-LM's 48.79\%. 
This is because grouping is ineffective when only one dataset is available, showing the importance of multi-LoRA for improved scheduling flexibility.
Second, as more adapters are trained together, the bubble ratio steadily decreases: 15.00\% for 2 adapters, 12.23\% for 3, and 11.09\% for 4. In comparison, mLoRA reaches only 34.11\%, confirming that \Name's grouping and batching significantly reduce pipeline idle time.
Lastly, with four adapters, the bubble ratio is 11.09\%. 
This is due to uneven execution times across pipeline stages, with the last stage taking longer because it handles an extra linear layer and cross-entropy loss.
This limitation is not solvable by our scheduler and is out of scope for this work.

\begin{figure}[t]
    \centering
    \includegraphics[width=0.99\linewidth]{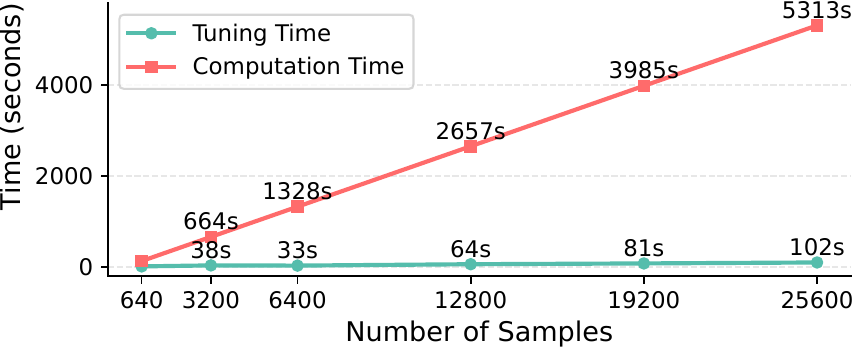}
    \caption{Tuning and computation time vs. number of samples for 4-stage pipeline with 4 adapters.}
    \label{fig:evaluation-fig-9-tuning-time}
\end{figure}

\myparagraph{Tuning Time} 
Figure~\ref{fig:evaluation-fig-9-tuning-time} shows how tuning and computation time grow with the number of training samples for a 4-stage pipeline with 4 adapters, measured on 64 vCPUs and 4 H100 GPUs.
The scheduling time increases nearly linearly, from 15.74 seconds at 640 samples to 102.12 seconds at 25600 samples, demonstrating linear scalability of our scheduler.
The computation time also increases nearly linearly, with a much larger slope than the scheduling time.
The scheduling overhead is negligible for three reasons.
First, the CPU-based scheduling runs in parallel with GPU training of the preceding global batch, with linear scaling of ~4ms per sample in CPU and magnitude difference in execution time between CPU and GPU, making the scheduler's latency fully hidden by this overlap.
Second, as shown in Figure~\ref{fig:evaluation-fig-8-pipeline-bubbles}, the performance gains saturate at 4 adapters, allowing practical deployment with a small constant number of adapters.
Third, we implement a timeout on the MILP solver and fall back to a greedy bin-packing algorithm if the solver takes too long, allowing us to configure the scheduler to balance effectiveness and efficiency, ensuring the scheduling overhead is always within a controllable range.

\myparagraph{Effectiveness of the Merging \& Greedy Fallback}
We evaluate the effectiveness of our scheduler's merging and greedy fallback components on 4 adapters of LLaMa-3.1-70B fine-tuned on four H100 GPUs.
The merging pass improves throughput by 4.34\%, while the two-stage MILP optimization provides an additional 3.82\% improvement over pure greedy bin-packing.
The MILP solver path is selected for 77.4\% of global batches with a timeout of 10 seconds, indicating its effectiveness in reducing token counts for underfilled microbatches.
These modest improvements reflect that most microbatches are already well-packed, and our algorithms primarily optimize the final microbatch in each global batch.
Since scheduling overhead is hidden by parallel GPU execution, these optimizations push performance toward the hardware limit without introducing additional overhead.

\subsection{Speedup Breakdown}
\label{sec:evaluation-speedup-breakdown}

\begin{figure}[t]
    \centering
    \includegraphics[width=0.99\linewidth]{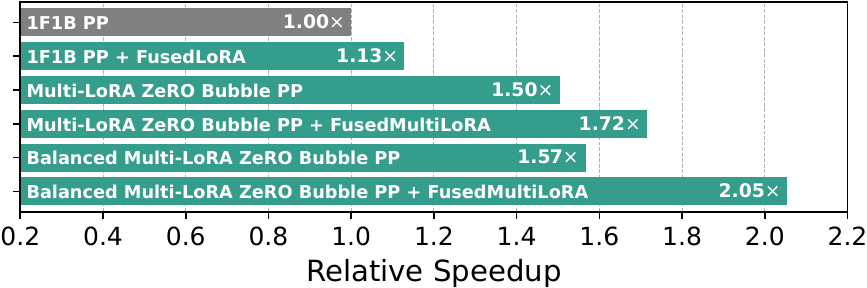}
    \caption{Speedup breakdown of \Name on LLaMa-3.1-70B with 4 GPUs.}
    \label{fig:evaluation-fig-4-speedup-breakdown}
\end{figure}

Figure~\ref{fig:evaluation-fig-4-speedup-breakdown} shows the contribution of each component in \Name. 
Starting from the baseline one forward one backward (1F1B) pipeline parallelism used in Megatron-LM, adding FusedLoRA alone yields a 1.13$\times$ speedup. 
This modest gain is constrained by load imbalance and suboptimal token shapes, which limit kernel efficiency (see Figure~\ref{fig:evaluation-fig-5-kernel-perf-fwd_bwd}). 
Replacing 1F1B with Multi-LoRA zero-bubble pipeline parallelism improves throughput to 1.50$\times$ by eliminating pipeline stalls through more microbatches from independent adapters. 
Adding FusedMultiLoRA kernel further raises the speedup to 1.72$\times$ by enabling multi-adapter microbatches and reducing redundant memory access. 
When we apply our scheduler to rebalance token distribution across microbatches, performance improves to 1.57$\times$ even without fusion, as it significantly reduces load imbalance. 
Finally, combining adaptive scheduling with fused kernels achieves the highest speedup of 2.05$\times$, showing the importance of jointly optimizing kernel efficiency, parallelism, and workload balance.

The speedup over mLoRA is driven by two main optimizations: (i) our kernel fusion yields a 1.15$\times$ speedup, as seen by comparing bars 3 and 4 in Figure~\ref{fig:evaluation-fig-4-speedup-breakdown}, with even greater gains when the sequence length is regular and matches the performant sequence length of our kernel (bars 5 and 6); and (ii) our adaptive batching mitigates load imbalance, providing a 1.19$\times$ speedup (bars 4 and 6). These improvements are further supported by microbenchmarks in Figure~\ref{fig:evaluation-fig-5-kernel-perf-fwd_bwd} (1.17$\times$ average speedup, up to 1.24$\times$ for kernel performance) and Figure~\ref{fig:evaluation-fig-8-pipeline-bubbles} (23.02\% reduction in pipeline bubbles).

\section{Discussion and Future Work}
\label{sec:discussion}

\myparagraph{Generalizability to LoRA Variants}
Our kernel fusion design is extensible to other popular LoRA variants like DoRA~\cite{liu2024dora} and VeRA~\cite{kopiczko2023vera}.
These methods typically add pre- or post-processing functions around the core LoRA computation.
Our optimizations are orthogonal to these modifications, and users can define prologue/epilogue functions to extend our kernels.
While manual extension is effective, a more general approach is to integrate our fusion patterns into a compiler framework.
As future work, we plan to leverage \texttt{torch.compile} by adding compiler annotations as hints that guide the fusion process of the LoRA pattern.
This would automate the optimization for both existing and future LoRA variants, eliminating the need for manual kernel development and system expertise from users.

\myparagraph{Generalizability to Quantization}
The kernels proposed in \Name can be directly applied to 4-bit QLoRA~\cite{dettmers2023qlora}. 
Current QLoRA implementations dequantize 4-bit weights to half-precision before LoRA computation, allowing our kernels to work without modification.
While dequantization could be fused with the LoRA path, recent work shows that two-step approaches are often more performant for large token counts~\cite{ding2025tilus}.

\section{Related Work}
\label{sec:related-work}

\sloppy

\myparagraph{System-level Optimizations on LoRA}
While there has been extensive work at the algorithmic level to make fine-tuning more stable and efficient, system-level optimizations for LoRA are mostly for inference and serving. 
PetS~\cite{zhou2022pets} is the first work that proposes multi-task parameter-efficient fine-tuned transformers serving and introduces a scheduling algorithm to coordinate different requests. Punica~\cite{chen2024punica}, S-LoRA~\cite{sheng2023s-lora}, and dLoRA~\cite{wu2024dlora} propose serving multiple LoRA adapters together to increase system throughput.
For LoRA fine-tuning, research interest is more focused on privacy preserving.
Offsite-tuning~\cite{xiao2023offsite} and DLoRA~\cite{gao2024Dlora-privacy} propose to decouple the large model owner and the data owner and connect them with lightweight adapters to enhance privacy.
In concurrent work, LobRA~\cite{lin2025lobra}, explored addressing multi-tenant fine-tuning over heterogeneous data. Its scheduling approach is complementary to our contributions. Our kernel fusion provides an orthogonal optimization that can directly enhance LobRA's performance, while our pipeline-aware scheduler could further reduce pipeline bubbles within LobRA as an additional improvement.

\myparagraph{Model Batching in Training}
Model batching improves hardware utilization by co-scheduling multiple training jobs on shared hardware. 
AutoML frameworks~\cite{feurer-neurips15a,feurer-arxiv20a} and TUPAQ~\cite{sparks2015automating} train numerous candidate models in parallel for architecture search, while Ease.ml~\cite{li2018easeml} focuses on multi-tenant model-selection. 
HFTA~\cite{wang2021hfta} proposes to horizontally fuse models from repetitive jobs at the operator level for better hardware utilization.
Multi-tenancy has also been applied to federated learning~\cite{zhuang2022smart-multi-tenant-federated-learning}. 
Unlike these general approaches, which are not model-aware, \Name is tailored for multi-tenant LoRA fine-tuning. 
It exploits the shared base model to reduce memory usage and address distributed training bottlenecks like pipeline bubbles and load imbalance.

\myparagraph{Kernel Fusion}
Kernel fusion is widely used to reduce redundant memory access and improve performance.
Compiler-based approaches such as TVM~\cite{tvm}, XLA~\cite{xla}, Ansor~\cite{zheng2020ansor}, TensorIR~\cite{feng2023tensorir}, torch.compile~\cite{ansel2024pytorch}, and Hidet~\cite{ding2023hidet} focus on automatic fusion via scheduling or tuning. Graph-level optimization frameworks like TASO~\cite{jia2019taso}, PET~\cite{wang2021pet}, and Automatic Horizontal Fusion~\cite{li2022automatic} perform rule- or cost-based transformations to eliminate redundant computation and improve fusion opportunities. Inference-oriented systems such as DNNFusion~\cite{niu2021dnnfusion}, ASPEN~\cite{park2023aspen}, AStitch~\cite{zheng2022astitch}, TensorRT~\cite{tensorrt}, ONNXRuntime~\cite{onnxruntime}, Rammer~\cite{ma2020rammer}, and Roller~\cite{zhu2022roller} use various fusion strategies to optimize runtime performance. Manual approaches like Triton~\cite{tillet2019triton} let developers implement custom fused kernels with fine-grained control.
Recently, Mirage~\cite{wu2024mirage} introduces a multi-level superoptimizer that automatically fuses complex tensor program blocks and shows its benefits for LoRA serving. However, it does not yet address LoRA fine-tuning scenarios with a sufficient number of tokens, dropout, backward computation, or fusion challenges from multi-LoRA kernel execution.

\myparagraph{Parallelism and Distributed Training}
A lot of work has been done on parallelizing the training of large models, such as Data Parallelism~\cite{pytorch-distributed,tensorflow}, Sharded Data Parallelism~\cite{pytorch-fsdp,zero,zero-offload,zero-infinity}, Tensor Parallelism~\cite{megatron-1}, and Pipeline Parallelism~\cite{pipedream,gpipe,dapple}.
Hybrid parallelism is usually used to combine multiple parallelism strategies to achieve better performance~\cite{megatron-1,zheng2022alpa,boehm2014hybrid}.
To effectively find the optimal parallelization strategies, systems~\cite{zheng2022alpa,flexflow,unger2022unity,liu2024aceso,lin2024nnscaler,miao2023galvatron,zhu2025mist} are proposed to automatically find the best combination of parallelism. These automatic planners are orthogonal to our work because any parallelization strategy they produce can directly benefit from our fused kernels.

\section{Conclusion}
\label{sec:conclusion}

This paper identifies and addresses two critical performance bottlenecks in LLM LoRA fine-tuning: 
redundant memory access in LoRA modules and missed optimization opportunities for grouping multiple concurrent LoRA jobs. 
Our solution, \Name, introduces a novel horizontal fusion technique tailored for LoRA kernels that reduces memory traffic by up to 37\% and a complementary job-level scheduling strategy that improves GPU utilization from 65\% to 89\%. 
Combined, these optimizations achieve up to 1.96$\times$ speedup compared to the state-of-the-art systems across various models and datasets. 
We hope \Name will help improve the accessibility and efficiency of LLM LoRA fine-tuning for both researchers and practitioners.

\section{Acknowledgement}
We sincerely thank our shepherd, Matthias Boehm, and the anonymous reviewers for their valuable feedback.
We also appreciate members of the EcoSystem Research Laboratory at the University of Toronto for their discussions and suggestions, with special thanks to Yu Bo Gao and Xiao Zhang for their contributions.
The authors with the University of Toronto are supported by Vector Institute Research grants, the Canada Foundation for Innovation JELF grant, NSERC Discovery grant, AWS Machine Learning Research Award (MLRA), Facebook Faculty Research Award, Google Scholar Research Award, and VMware Early Career Faculty Grant.

%

\appendix
\section{Artifact Appendix} 

\subsection{Abstract}

We provide the source code of \Name and scripts to reproduce the major experimental results from the  paper.
The artifact enables reproduction of key evaluation figures including data distribution analysis (Figure~\ref{fig:evaluation-fig-0-data-statistics}), end-to-end performance comparisons (Figure~\ref{fig:evaluation-fig-1-end-to-end}), kernel-level performance analysis (Figure~\ref{fig:evaluation-fig-5-kernel-perf-fwd_bwd}), layer-wise performance evaluation (Figure~\ref{fig:evaluation-fig-6-layer-perf-normalized}), and memory traffic analysis (Figure~\ref{fig:evaluation-fig-7-kernel-ncu-profile}).
The artifact includes detailed installation procedures and automated evaluation workflows.
Full reproduction requires a Linux system with 192 GB RAM, 256 GB disk space, and 4 NVIDIA H100 GPUs with NVLink interconnects.
Kernel and layer-level benchmarks can be executed on systems with a single GPU.

\subsection{Description \& Requirements}


\subsubsection{How to access}
The code is available at: 
Github \url{https://github.com/CentML/lorafusion} and Zenodo \url{https://zenodo.org/records/17051801}.

\subsubsection{Hardware dependencies}
The complete experimental evaluation requires a Linux system equipped with at least 192 GB of system memory, 256 GB of available disk storage, and 4 NVIDIA H100 GPUs interconnected via NVLink.

For partial evaluation, kernel and layer-level benchmarks can be executed on systems with a single GPU. Performance results on alternative hardware configurations may differ from those reported in the paper. Systems with higher compute-to-memory bandwidth ratios typically yield superior performance, while older hardware with lower ratios may exhibit reduced performance gains.

\subsubsection{Software dependencies}
The artifact requires Conda for environment management. The software stack includes CUDA 12.4, PyTorch v2.6.0, megatron-core v0.11.0, and Triton v3.2.0. All dependencies are automatically installed through the provided setup scripts.

\subsubsection{Benchmarks} 
None

\subsection{Set-up}

\begin{enumerate}
    \item Clone the GitHub repository.
    \vspace{3pt}
\begin{lstlisting}[language=bash]
git clone https://github.com/CentML/lorafusion.git
cd lorafusion
git checkout eurosys-ae
\end{lstlisting}

    \item Install the requirements by running this command or following \texttt{docs/installation.md}.
    \vspace{3pt}
\begin{lstlisting}[language=bash]
conda create -y -n lorafusion python=3.12
conda activate lorafusion
cd benchmarks_paper
bash scripts/setup/setup_env.sh
\end{lstlisting}

    \item Download the Hugging Face models and datasets. Make sure you are logged in and have access to them.
    \vspace{3pt}
\begin{lstlisting}[language=bash]
# huggingface-cli login
python prepare_models.py
python gen_sample_distribution.py
\end{lstlisting}

    \item Verify hardware-specific kernel configurations. The Triton kernels require hardware-specific tuning to optimize tiling strategies. Examine \texttt{lorafusion/ops/triton\_ops/config.py} to determine if pre-tuned configurations exist for the target hardware. Pre-configured settings are available for:
    \begin{itemize}
        \item NVIDIA H100 80GB HBM3 (recommended)
        \item NVIDIA A100 SXM4 80GB
        \item NVIDIA A100 PCIe 80GB
        \item NVIDIA GeForce RTX 3090
    \end{itemize}
    For unsupported hardware configurations, execute the kernel tuning process:
    \vspace{3pt}
\begin{lstlisting}[language=bash]
cd /PATH/TO/lorafusion/
python tools/tune_kernels.py
\end{lstlisting}
    Then, update \texttt{lorafusion/ops/triton\_ops/config.py} with the generated optimal configurations.
\end{enumerate}

\subsection{Evaluation workflow}

\subsubsection{Major Claims}

\begin{itemize}
    \item (C1): \Name is up to $1.96\times$ faster (average $1.47\times$) than Megatron-LM, and up to $1.46\times$ faster (average $1.29\times$) than mLoRA. See Section~\ref{sec:evaluation-end-to-end-results} and Figure~\ref{fig:evaluation-fig-1-end-to-end}.
    \item (C2): Our fused kernels are up to $1.39\times$ faster (average $1.27\times$) and can replace existing LoRA kernels. See Section~\ref{sec:evaluation-effectiveness-of-fused-lora-kernel} and Figure~\ref{fig:evaluation-fig-5-kernel-perf-fwd_bwd}, Figure~\ref{fig:evaluation-fig-6-layer-perf-normalized}, and Figure~\ref{fig:evaluation-fig-7-kernel-ncu-profile}.
\end{itemize}

\subsubsection{Complete Experimental Evaluation}

The full experimental evaluation requires 4 NVIDIA GPUs, each with 80GB memory capacity.

\begin{enumerate}
    \item Navigate to the \texttt{benchmarks\_paper} directory.
    \item Execute the complete evaluation suite:
    \vspace{3pt}
\begin{lstlisting}[language=bash]
bash scripts/run_all.sh all
\end{lstlisting}
    \begin{enumerate}
        \item The complete evaluation encompasses all primary experiments and kernel performance assessments, requiring approximately 4 hours of computation time.
        \item Detailed command specifications and timing estimates are available in \texttt{scripts/run\_all.sh}.
        \item Individual experiment subsets can be executed by modifying the script parameters.
    \end{enumerate}
    \item Evaluation results are generated in the \texttt{results} directory, producing figures corresponding to Figure~\ref{fig:evaluation-fig-0-data-statistics}, Figure~\ref{fig:evaluation-fig-1-end-to-end}, Figure~\ref{fig:evaluation-fig-5-kernel-perf-fwd_bwd}, Figure~\ref{fig:evaluation-fig-6-layer-perf-normalized}, and Figure~\ref{fig:evaluation-fig-7-kernel-ncu-profile}.
\end{enumerate}

\subsubsection{Reduced-Scale Evaluation}

For systems with limited GPU resources, kernel and layer-level benchmarks can be executed on a single GPU configuration.

\begin{itemize}
    \item Systems with GPUs containing 80GB or greater memory capacity can execute the comprehensive single-GPU evaluation suite:
    \vspace{3pt}
\begin{lstlisting}[language=bash]
bash scripts/run_all.sh all_single_gpu
\end{lstlisting}
    \item Systems with GPUs containing less than 80GB memory can execute kernel and layer benchmarks independently:
    \vspace{3pt}
\begin{lstlisting}[language=bash]
bash scripts/run_all.sh layer
bash scripts/run_all.sh kernel
\end{lstlisting}
\end{itemize}

Results are generated in the \texttt{results} directory. For 80GB+ configurations, the evaluation produces a subset of Figure~\ref{fig:evaluation-fig-1-end-to-end} alongside Figure~\ref{fig:evaluation-fig-5-kernel-perf-fwd_bwd}, Figure~\ref{fig:evaluation-fig-6-layer-perf-normalized}, and potentially Figure~\ref{fig:evaluation-fig-7-kernel-ncu-profile} (contingent on NCU profiling availability).

\subsection{Notes on Reusability}

Experimental customization can be achieved by modifying \texttt{scripts/run\_all.sh} and associated sub-scripts. The artifact provides evaluation scripts and corresponding visualization tools for result generation.

Performance characteristics on alternative GPU architectures may differ from H100-based results. Systems with lower compute-to-memory bandwidth ratios typically exhibit reduced performance gains. Power consumption constraints during kernel configuration tuning may affect optimal parameter selection and subsequent benchmark accuracy. For consistent performance evaluation across different hardware, manual GPU frequency configuration is recommended:
\vspace{3pt}
\begin{lstlisting}[language=bash]
# Disable automatic frequency scaling
sudo nvidia-smi -pm 1
sudo nvidia-smi --auto-boost-default=0

# Query supported frequency configurations
nvidia-smi -q -d SUPPORTED_CLOCKS

# Configure specific memory and graphics clock frequencies
# sudo nvidia-smi -ac <memory_clock,graphics_clock>
# e.g.,
# sudo nvidia-smi -ac 6251,1050
\end{lstlisting}




\bibliographystyle{ACM-Reference-Format}
\bibliography{references}

\end{document}